\documentclass[a4paper,12pt]{article}

\usepackage{url}
\usepackage[numbers]{natbib}
\usepackage[dvipsnames]{xcolor}
\usepackage[utf8]{inputenc} 
\usepackage[T1]{fontenc}
\usepackage[a4paper, total={7in, 9in}]{geometry}
\usepackage{graphicx}
\graphicspath{{figures/}} 
\usepackage{amscd,amsmath,amssymb,amsfonts,latexsym,mathrsfs,amsthm,mathtools}				
\usepackage{bm}                         
\usepackage{bbm}
\usepackage{isomath}                    
\usepackage{subcaption}
\usepackage{multirow} 
\usepackage{hyperref}

\usepackage{newfloat}
\DeclareFloatingEnvironment[fileext=lof,listname={List of algorithms}, 
                            name=Algorithm, placement=h]{algorithm}

\makeatletter
\newcommand{\notprop}{\propto\kern-1\@ptsize pt \diagup}
\makeatother

\newcommand{\Exp}{\mathbb{E}} 
\newcommand{\Eq}[1]{Eq.~{\eqref{#1}}}

\newcommand{\R}{\mathbb{R}}
\newcommand{\vtau}{{\boldsymbol\tau}}
\newcommand{\vtheta}{{\boldsymbol\theta}}
\newcommand{\vxi}{{\boldsymbol\xi}}

\newcommand{\vpsi}{{\boldsymbol\psi}}
\newcommand{\y}{{\bf y}}

\newcommand{\Y}{{\bf Y}}
\newcommand{\z}{{\bf z}}

\newcommand{\fapprox}{\widetilde{f}}

\definecolor{myblue}{RGB}{0,163,243}

\usepackage{soul}

\usepackage[normalem]{ulem}

\newtheorem{rem}{Remark}

\begin{document}

\title{
 A survey of Monte Carlo methods for noisy and costly densities
with application to reinforcement learning and ABC
}

\author{F. Llorente$^{\ddagger}$, L. Martino$^{\star}$, J. Read$^\dagger$, D. Delgado$^*$  \\
	{\footnotesize$^{\ddagger}$  Stony Brook University,  Stony Brook (USA).}\\
	{\footnotesize$^\star$  Universit{\'a} degli Studi di Catania,  Catania (Italy).} \\
	{\footnotesize$^\dagger$ École Polytechnique, Palaiseau (France).}	\\
	{\footnotesize$^*$  Universidad Carlos III de Madrid,  Legan\'es (Spain).}
}

\maketitle

\begin{abstract}
	This survey gives an overview of Monte Carlo methodologies using surrogate models, for dealing with densities which are intractable, costly, and/or noisy. This type of problem can be found in numerous real-world scenarios, including stochastic optimization and reinforcement learning, where each evaluation of a density function may incur some computationally-expensive or even physical (real-world activity) cost, likely to give different results each time. The surrogate model does not incur this cost, but there are important trade-offs and considerations involved in the choice and design of such methodologies. We classify the different methodologies into three main classes and describe specific instances of algorithms under a unified notation. A modular scheme which encompasses the considered methods is also presented. A range of application scenarios is discussed, with special attention to the likelihood-free setting and reinforcement learning. Several numerical comparisons are also provided. 
\end{abstract}

\section{Introduction}
\label{sec:intro}

Bayesian methods and their implementations by means of sophisticated Monte Carlo techniques, such as Markov chain Monte Carlo (MCMC) and importance sampling (IS) schemes, have become very popular \cite{Robert04,liu2008monte}. 
In the last years, there is a broad interest in performing Bayesian inference in models where the posterior probability density function (pdf) is analytically {\it intractable}, and/or {\it costly} to evaluate, and/or its evaluation is {\it noisy} \cite{arnold2012noisy,jin2005evolutionary,nissen1998optimization}. Namely, there are several practical situations where the posterior distribution cannot be evaluated pointwise or its evaluation is expensive \cite{everitt2015bayesian,alquier2016noisy,park2018bayesian,luengo2020survey}.  Such models occur in a wide range of applications including spatial statistics, social network analysis, statistical genetics, finance, etc.
For instance, {\bf (a)} for the use of  massive  datasets where the likelihood consists of a product of a large number of terms \cite{bardenet2017markov}, or  {\bf (b)} for the existence of a large number of latent variables that we should marginalize out (hence, the posterior pdf can be obtained only solving a high dimensional integral) \cite{andrieu2009pseudo}.  Moreover, another scenario is {\bf (c)} when a piece of likelihood  function is analytically unknown and it should be approximated \cite{park2018bayesian,everitt2015bayesian}. The intractable likelihood models arising from, for example, Markov random fields, such as those found in spatial statistics and network analysis \cite{rue2005gaussian}.
In many settings, {\bf (d)} the likelihood function is induced by a complex stochastic computer model which is costly to evaluate pointwise \cite{llorente2021deep}.  {\bf (e)} In other application fields, such as reinforcement learning, a target function (usually a policy) cannot be exactly evaluated neither quickly nor precisely, since such an evaluation corresponds to interaction with an environment (possibly in the real world) which is inherently lengthy to obtain and susceptible to contamination by noise perturbation. Hence, the evaluation is obtained with a certain degree of uncertainty \cite{cleary2021calibrate}.
 \newline 
{\bf Noisy computational schemes.} The solutions proposed in the literature to performing the inference in the scenarios {\bf (a)}-{\bf (b)}-{\bf (c)} above, have been carried out using Monte Carlo algorithms which often consider noisy evaluations of the target density \cite{bardenet2017markov,andrieu2009pseudo,PMCMC,park2018bayesian}.  A natural approach in these cases is to replace the intractable/costly model with an approximation (or with a pointwise estimation in the case of a noisy model). Thus, the corresponding Monte Carlo schemes also involve the use a {\it surrogate model} via regression techniques. 
Furthermore, in the scenario {\bf (d)},  if it is possible to draw artificial data according the observation model, sometimes is preferable to generate fake data (given some parameters) and to measure the discrepancy between the generated data and the actual data, instead of  evaluating the costly likelihood function \cite{beaumont2019approximate,luengo2020survey}. This approach is known as Approximate Bayesian Computation (ABC). This area has generated much activity in the literature (see, e.g., \cite{robert2016approximate}).  The  discrepancy measure plays the role a surrogate model and, due to the stochastic generation of the artificial data, it also adds uncertainty (i.e., as a noise perturbation) in the internal evaluations within the ABC-Monte Carlo methods \cite{luengo2020survey}. Finally, The last scenario {\bf (e)} is intrinsically noisy, so that it also requires specific computational solutions.   

The three different cases above, {\it intractable}, {\it costly} and {\it noisy} evaluations of a posterior distribution can appear and/or can be addressed separately \cite{llorente2021deep,alquier2016noisy,luengo2020survey}. In all of these cases, a surrogate model can accelerate the Monte Carlo method or approximate the posterior distribution \cite{conrad2016accelerating,park2020function,sherlock2017adaptive,jarvenpaa2021parallel}. As described above,  these cases also appear jointly in real-world applications (specially, if we consider the algorithms designed to address those issues): `intractable and costly',  `intractable and noisy', or `costly and noisy' posterior evaluations, etc.  The challenge posed by these contexts has led to the development of recent theoretical and methodological advances in the literature. Furthermore, surrogate models have been considered as an alternative to Monte Carlo for approximating complicated integrals. Here, the surrogate is substituted {\it directly} into the integral of interest, instead of the original density (e.g., a posterior). A cubature rule is subsequently obtained, which makes a more efficient use of the posterior evaluations \cite{briol2019probabilistic,kanagawa2019convergence,karvonen2018bayes}. 
\newline
{\bf Contribution.} In this work, we provide a survey of methods which use surrogate models {\it within} Monte Carlo algorithms for dealing with noisy and costly posteriors. Some of them have been introduced only in the context of expensive posteriors \cite{conrad2016accelerating}. Other schemes have been designed only for improving the efficiency of the Monte Carlo methods considering a more sophisticated proposal density (see for instance, \cite{martino2018adaptive,llorente2021deep}). However, all of them can be applied also in a noisy scenario. 
In Sections  \ref{GenFrame} and \ref{GenScheme}, we provide a general joint framework which encompasses most of the techniques in the literature.  We introduce the vanilla schemes for noisy MH method (well studied in the literature, e.g., \cite{andrieu2009pseudo,doucet2015efficient}) and also of a noisy IS scheme (which also has been studied in the literature in works such as \cite{fearnhead2010random,tran2013importance}).  We focus mainly on the static batch scenario for MCMC and IS algorithms. However, most of the results presented in this work can be extended to the sequential framework (consider, e.g., the recent work of \cite{bon2021accelerating}).
\newline
In Section \ref{GenFrame}, we start describing the general framework.
In Section \ref{GenScheme}, we classify the studied techniques in different families, and  provide several explanatory tables and figures.  More specifically, we divide the algorithms in the literature in three broad classes: 1) two-stage, 2) iterative refinement,  and 3) exact. 
In Section \ref{SectionSpecificExample}, we also provide detailed descriptions of specific examples of algorithms. For instance, we provide a generic description of Metropolis-Hastings (MH) schemes on an iterative surrogate. The {\it moving target MH} algorithm is a specific example of this \cite{ying2020moving}.  Then, we describe some specific implementation of the so-called {\it Delayed Acceptance MH}  (DA-MH) methods \cite{banterle2019accelerating}. We also introduce {\it Noisy Deep Importance Sampling} (N-DIS) which is a noisy version of the Deep IS method in \cite{llorente2021deep}.  { Section \ref{TradeoffSect} is devoted to some theoretical discussions regarding the choice of the parameters.} 
The range of application of those methods is firstly discussed in Section \ref{sec_appEsc}. 
Furthermore, more specifically, we give a detailed description of two scenarios: the likelihood-free approach in Section \ref{sec_abc}, and the reinforcement learning (RL) setting in Section \ref{RLsect} \cite{tavakol2018markov,hoffman2008trans}. 
We test the presented algorithms in different numerical experiments in Section \ref{numexperimentSect}.
The application to a benchmark RL problem, the double cart-pole system \cite{heidrich2009neuroevolution}, is given in Section \ref{Double_cart_poleSect}. Finally, we conclude with a brief discussion in Section \ref{sec_conc}.

\section{General framework}\label{GenFrame}

Let us assume that our goal is the study of the unnormalized density $p(\vtheta)$, where $ \vtheta\in\bm{\Theta}\subset \mathbb{R}^d$ is a vector of parameters of interest, using Monte Carlo methods. {  In many applications, the direct study of $p(\vtheta)$ is precluded by the impossibility of evaluating it at any $\vtheta$ (see section below).  
}
There are two related problems: \textbf{(P1-noisy)} for any $\vtheta$, we cannot evaluate $p(\vtheta)$ exactly, but  we only have access to a related noisy realization, and  \textbf{(P2-costly)} { evaluating $p(\vtheta)$ (or obtaining such a noisy realization) is very expensive}.
Typically, this occurs in applications where the density of interest $p(\vtheta)$ is intractable or expensive to evaluate. 
{ Both scenarios, noisy (P1) and costly (P2), can be addressed by applying  surrogate models (approximating $p(\vtheta)$) and using them within Monte Carlo schemes. Hence, the solutions described in the rest of this work can be applied in both cases. Below, we introduce and describe the notation of the more sophisticated case from a practical and theoretical point of view, i.e., the noisy scenario. }
\newline
\newline
{ {\bf Noisy scenario.}} More specifically,  in many practical cases, we have access to a noisy realization $\widetilde{p}_M(\vtheta)$ related to $p(\vtheta)$.
{Namely, for a given $\vtheta$, $\widetilde{p}_M(\vtheta)$ with $M\in \mathbb{N}$ is a random variable (more precisely an estimator, where generally $M$ represents the number of used samples) with
\begin{equation}
\mathbb{E}[\widetilde{p}_M(\vtheta)] =m(\vtheta)=p(\vtheta)-\mu_M(\vtheta),\quad  \texttt{var}[\widetilde{p}_M(\vtheta)] = s_M^2(\vtheta),
\end{equation}
for some {\em mean function}, $m(\vtheta)=p(\vtheta)-\mu_M(\vtheta)$ where $\mu_M(\vtheta)$ is a bias, and {\em variance function}, $s_M^2(\vtheta)$. 
The bias $\mu_M(\vtheta)$ can also depend on other (non-random) parameters, for instance $\mu_M(\vtheta,\epsilon)$, as shown in the specific application in Section \ref{sec_abc}.
The integer parameter $M$ and the  parameter $\epsilon$ are such that 
\begin{align}
\mu_M(\vtheta,\epsilon) \rightarrow 0,  \quad s_M^2(\vtheta) \rightarrow 0, 
\end{align}
when $M\rightarrow \infty$ (and, e.g., $\epsilon\rightarrow 0$ in Section \ref{sec_abc}), for all $\vtheta\in \bm{\Theta}$.}
The unbiased case, {$\mu_M(\vtheta)=0$ for all $\vtheta$ hence $m(\vtheta)=p(\vtheta)$}, appears naturally in some applications, or it is often assumed as a pre-established condition by the authors.  
In some other scenarios, the noisy realizations are known to be unbiased estimates of some transformation of $p(\vtheta)$, e.g., of $\log p(\vtheta)$.  
{In this case, defining as $\tilde{\xi}(\vtheta)$ the estimation of $\log p(\vtheta)$, then $\widetilde{p}_M(\vtheta)=e^{\tilde{\xi}(\vtheta)}$ and  $\mathbb{E}[\widetilde{p}_M(\vtheta)] = p(\vtheta)-\mu_M(\vtheta)$ with a non-null bias, but  $\mu_M(\vtheta) \rightarrow 0$ as $M\rightarrow \infty$.  However, there are cases such as $\widetilde{p}_M( \vtheta) = \log p(\vtheta) + {u}$, { and the pdf $p_u(u)$ does not depend on $\vtheta$,} where we can take
$\widetilde{p}_M( \vtheta) = e^{\widetilde{\xi}(\vtheta)}$ which fulfills $\Exp[\widetilde{p}_M( \vtheta)] \propto p(\vtheta)$ \cite{jarvenpaa2021parallel,drovandi2018accelerating}.  Appendix \ref{EmuLogLikeSect} contains related material.
} 
\newline
\newline 
It is also possible to control the noise level {by varying the parameter $M$. In the different applications, this means increasing (a) the accuracy of auxiliary estimators including samples,} (b) adding/removing data to the mini-batches (e.g., in the context of Big Data) or (c) interacting with an environment over longer/shorter periods of time (e.g., in reinforcement learning). See also the illustrative example in Appendix \ref{illustrative_ex_1D} and the related Figure \ref{fig:toy1D}.
\newline 
{
{\bf Costly scenario.} Note that in the costly scenario (P2) we {may} have $s^2(\vtheta)=0$ and $\widetilde{p}_M( \vtheta)=p(\vtheta)$ for all $\vtheta$, but still we have the issue of the expensive evaluation of $p(\vtheta)$. Clearly, 
also in the noisy scenario, obtaining a realization of $\widetilde{p}_M( \vtheta)$ can be computational expensive. We refer to this case as `noisy and costly'' framework. We will see that the use of surrogate models within Monte Carlo methods are useful in all the cases, noisy or costly, and in the more general ``noisy and costly'' framework. }
\newline
\newline
{
A table of the main acronyms of the work is given in Table \ref{tab_acro} below.

\begin{table}[!h]
	\centering
	{
	\caption{{Table of the main acronyms of the work.} \label{tab_acro}}
	\vspace{-0.3cm}
	{\footnotesize
		\begin{tabular}{|c|l|}
			\hline
			MC & Monte Carlo \\ 
			MCMC & Markov Chain Monte Carlo \\ 
			MH & Metropolis-Hastings \\
			IS & Importance Sampling \\
			 N-MH &  Noisy Metropolis-Hastings \\
			 N-IS &  Noisy Importance Sampling \\
			 MCWM & Monte Carlo-within-Metropolis \\
			PM-MH & Pseudo Marginal Metropolis-Hastings \\
			DA-MH &	Delayed Acceptance Metropolis-Hastings\\
			DA-PM-MH &	Delayed Acceptance  Pseudo Marginal Metropolis-Hastings\\
			MH-S & Metropolis-Hastings on surrogate with iterative refinement	\\
			 DIS & Deep Importance Sampling \\
			 N-DIS & Noisy Deep Importance Sampling \\
			ABC & Approximate Bayesian Computation \\
			RL & Reinforcement Learning \\
			 GP &  Gaussian Process \\
			 k-NN &  k-Nearest Neighbors \\
			\hline
		\end{tabular}	
	}
	}
\end{table}

}

{
\subsection{Application scenarios}\label{sec_appEsc} 

In this section, a brief description of practical scenarios where we must handle noisy and costly target evaluations is provided.  In many of them, $p(\vtheta)$ may represent a posterior or a marginal-posterior density in a Bayesian inference problem. A list of different application frameworks is given below. However, regarding the approximate Bayesian computation (ABC) and reinforcement learning (RL), more details are provided in Sections \ref{sec_abc} and \ref{RLsect}, respectively.
\newline
\textbf{Pseudo-Marginal approach}: The unnormalized density $p(\vtheta)$ can be expressed as as an intractable integral over a nuisance variable ${\mathbf v}$. Namely, $p(\vtheta)$  is a marginal distribution, $p(\vtheta)=\int_{\mathcal{V}} p(\vtheta,{\mathbf v})d{\mathbf v}$   where  ${\mathbf v}$ is an auxiliary variable. More generally,
$$
p(\vtheta) \propto \int_{\mathcal{V}}  V(\vtheta,{\mathbf v}) p(\vtheta,{\mathbf v}) d{\mathbf v},
$$
where $p(\vtheta,\z)$ is a joint density and $V(\vtheta,\z)$ is non-negative integrable function.
Hence,  $p(\vtheta)$ cannot often be computed in closed-form.  This is the case of Bayesian models with auxiliary variables used, for instance, in the estimation of the volatility in state space models \cite{drovandi2018accelerating, lowe2023accelerating}. More generally, any model whose posterior has an intractable (marginal) likelihood function falls within this group \cite{ourRev}.
 When the aim is to run a MH algorithm on $p(\vtheta)$, rather than  on the joint $p(\vtheta,{\mathbf v})$, the evaluation of $p(\vtheta)$ at each $\vtheta$ can be estimated \textit{noisily} by using IS \cite{andrieu2009pseudo,andrieu2010particle} or sequential Monte Carlo (particle filters) \cite{drovandi2018accelerating, lowe2023accelerating}. Here, $\widetilde{p}_M( \vtheta)$ represents an approximation of the integral, i.e., an estimator of $p(\vtheta)$.
\newline
{\bf ABC, likelihood-free setting.} In the likelihood-free inference setting, it is assumed that the likelihood function is unknown or we cannot evaluate it, but we are able to generate independent data from it. In this scenario, substituting the intractable likelihood with an approximate likelihood is one possibility. This approximation is in turn  approximated pointwise with Monte Carlo using pseudo-data sets \cite{Marin12,price2018bayesian}. See Section \ref{sec_abc} for more details; $\widetilde{p}_M( \vtheta)$ is given in  Eq. \eqref{Mtheta_Eq}.
\newline
{\bf Doubly intractable posteriors.}  In certain statistical models such as exponential random graph models and Markov point processes \cite{park2018bayesian}, only a piece of the likelihood can be evaluated and another piece of the likelihood is unknown (typically a partition function $Z(\vtheta)=\int_{\mathcal{Y}} f(\y|\vtheta)d\y$ where $f(\y|\vtheta)$ is an {\it unnormalized} likelihood).
 This is the doubly intractable posterior setting. Note that differently from the ABC case, here some part of the likelihood is available. In this case, the unknown part of the likelihood must be estimated, so that the evaluation of the complete likelihood will be noisy. Given an estimator $\widehat{Z}(\vtheta)$ of $Z(\vtheta)$,  here $\widetilde{p}_M( \vtheta)=\frac{1}{\widehat{Z}(\vtheta)}  f(\y|\vtheta) g(\vtheta)$ (where $g(\vtheta)$ is a prior density), i.e., is a noisy version of $p(\vtheta)\propto \frac{1}{Z(\vtheta)}  f(\y|\vtheta) g(\vtheta)$.  
{
However, note that in general $\Exp[\widehat{Z}(\vtheta)^{-1}] \neq Z(\vtheta)^{-1}$ when $\widehat{Z}(\vtheta)$ is an unbiased estimate, hence using $\widetilde{p}_M( \vtheta)=\frac{1}{\widehat{Z}(\vtheta)}  f(\y|\vtheta) g(\vtheta)$ inside a MC algorithm will not approximate the posterior of interest (see noisy MH in the next section). The popular exchange algorithm \cite{murray2012mcmc} is an exact noisy MCMC algorithm that uses a one-sample approximation of the ratio $\frac{Z(\vtheta_{t-1})}{Z(\vtheta_\text{prop})}$ inside the acceptance probability. This algorithm is indeed a special case of the randomized MH (r-MH) algorithm presented in \cite{nicholls2012coupled} (see also next section). Observe also that there exist procedures for approximating directly the ratio $ \frac{1}{Z(\vtheta)}$: for instance, this is the case of the harmonic mean estimator or, more generally, the {\it reverse (a.k.a., reciprocal) importance sampling} method \cite{ourRev}. 
} 
 \newline
 {\bf Use of mini-batches (Big Data).}   There are cases where the likelihood is composed of a large product of terms, $\ell(\Y|\vtheta) = \prod_{i=1}^K \ell(\y_i|\vtheta)$ where $\Y$ is a matrix including the vector data $\y_i$ as (columns or rows). When there is a great amount of independent observations, the evaluation of the posterior,
$$
p(\vtheta) \propto g(\vtheta) \ell(\Y|\vtheta) = g(\vtheta)\exp\left( \sum_{i=1}^K \log\ell(\y_i|\vtheta) \right),
$$
can be extremely costly since it requires sweeping through all $K$ terms for the evaluation at any $\vtheta$. Namely, the evaluation of the likelihood function can be prohibitively expensive when there are huge amounts of data. In this context, a \textit{subsampling} strategy consists in computing the log-likelihood function using a random subset of data points, hence forming an unbiased estimator of the complete log-likelihood \cite{bardenet2017markov}. Namely, here we can define as $\widetilde{\xi}(\vtheta)=\frac{1}{k}\sum_{i=1}^k \log\ell(\y_{j_i}|\vtheta)$ where $k\leq K$ and $j_i\in \{1,...,K\}$, and $\widetilde{p}_M( \vtheta)=g(\vtheta)\exp [ K\widetilde{\xi}(\vtheta)]$.
{However, as we will see later, if we were to plug $\widetilde{p}_M( \vtheta)$ inside a MC algorithm, the algorithm actually approximates  $m(\vtheta) = g(\vtheta)\Exp\left[\exp [ K\widetilde{\xi}(\vtheta)]\right]$, which does not coincide with $p(\vtheta)$ since $\Exp\left[\exp [ K\widetilde{\xi}(\vtheta)]\right] \neq \ell(\Y|\vtheta)$.
Under the assumption that $\widetilde{\xi}$ is Gaussian distributed, one can account for this bias and correct the algorithms, which now fall into the framework by \cite{nicholls2012coupled} (see next section).  See \cite[Sect. 4]{bardenet2017markov} for a description of exact subsampling algorithms relying on a bias correction.}
Moreover, the subsampling estimates can used to run stochastic gradient-type Monte Carlo algorithms such as Langevin and Hamiltonian Monte Carlo \cite{bardenet2017markov}. 
 \newline
 {\bf Reinforcement learning (RL).} 
Direct policy search is an important branch of reinforcement learning, particularly in robotics \cite{deisenroth2013survey,handful}. In this context, $\vtheta$ is the parametrization of the policy of some agent, and $p(\vtheta)$ represents an expected return (i.e., a payoff function) for that policy. 
The expected return is approximated by the empirical return over an episode, i.e., the agent is run for a number of time steps and accumulates a payoff.
More details are given in Section \ref{RLsect}.
 \newline
{\bf Other application scenarios.} The topic of inference in noisy and costly settings is also of interest in the inverse problem literature, such as in the calibration of expensive computer codes \cite{duncan2021ensemble,cleary2021calibrate,bliznyuk2008bayesian}. 
Noisy likelihood evaluations are also considered for building surrogates, and then use them in order to obtain a variational approximations to the posterior \cite{acerbi2020variational}. 
}

\subsection{Vanilla schemes for Noisy MH and noisy IS} \label{VanillaSchemesSect}

In this Section, we present two basic Monte Carlo algorithms working with noisy realizations $\widetilde{p}_M( \vtheta)$. 
{ For the sake of simplicity, we assume $\widetilde{p}_M( \vtheta)$ is always non-negative.}
\newline
\textbf{Noisy MH.} The standard MH algorithm produces correlated samples from a target distribution $p(\vtheta)$ by  sampling candidates from a proposal density which are either rejected or accepted according to a suitable probability. The evaluation of the target density $p(\vtheta)$ is required at each iteration. A noisy version of this algorithm is obtained when we substitute the evaluations of $p(\vtheta)$ (at the candidate points) with a realization of the random variable $\widetilde{p}_M( \vtheta)$. The algorithm is shown in Algorithm \ref{tab_pseudoMH}. 
If a different noisy realization $\widetilde{p}_M( \vtheta_{t-1})$ is obtained at each iteration, this algorithm is called {\it Monte Carlo-within-Metropolis} (MCWM)\cite{andrieu2009pseudo,medina2016stability}. On the contrary, if it is recycled from the previous iteration, the algorithm is called {\it pseudo-marginal} MH (PM-MH) algorithm \cite{beaumont2003estimation,andrieu2009pseudo}. The latter approach ensures the algorithm is ``exact'',\footnote{{Although the correct term would be ``approximate-exact'', throughout this work we say ``exact'' to indicate when a Monte Carlo algorithm targets the desired distribution, {$m(\vtheta)$, in the case of a non-zero bias, or $p(\vtheta)$ when $\mu_M(\vtheta)=0$}.}} in the sense of having (the density proportional to) $m(\vtheta)=p(\vtheta)-\mu_M(\vtheta)$ as stationary distribution and the estimators derived from it converging to expectations under $m(\vtheta)$. 
Namely, as the number of iteration $T$ grows, we have the convergence of the chain to $m(\vtheta)$. 
{Note that, while PM-MH is commonly applied when $\mu_M(\vtheta)=0$, and hence the convergence is towards the $p(\vtheta)$ of interest, the algorithm works as long as the random realizations are non-negative and have finite expected value, that we generally denote as $m(\vtheta)$.}
Recall that even in the biased case, when increasing $M$, $m(\vtheta)\rightarrow p(\vtheta)$.  
\newline
{
It is worth mentioning here another instance of MH with a randomized acceptance probability studied in \cite{nicholls2012coupled}, called {\it randomized} MH (r-MH), that is complementary to the PM-MH approach. Specifically, the authors show the case of a MH with a random acceptance probability having correction terms to ensure the convergence with respect to the right distribution. This r-MH algorithm contains as special cases some (exact) algorithms proposed in the literature to deal with intractable/costly densities such as the \cite{murray2012mcmc} for doubly-intractable posteriors, or \cite{quiroz2018speedingJASA} for data-subsampling approximations. In most cases, the r-MH algorithm is not realizable in practice since the correction terms cannot be computed. The scheme is used by the authors to show that, under some conditions, the MCWM algorithm produces the same sequence of samples as an exact r-MH, out of the first $\mathcal{O}(M)$ MCMC samples, where $M$ is the number of auxiliary samples used to estimate the log-ratio $\log\frac{p(\vtheta_\text{prop})}{p(\vtheta_{t-1})}$}.
\newline
\textbf{Noisy IS.} 
In a standard IS scheme, a set of samples is drawn from a proposal density $q(\vtheta)$. Then each sample is weighted according to the ratio $\frac{p(\vtheta)}{q(\vtheta)}$. 
Like in the MH case, a noisy version of importance sampling can be obtained when we substitute the evaluations of $p(\vtheta)$ with noisy realizations of $\widetilde{p}_M( \vtheta)$.
See Table \ref{tab_noisyIS}. 
{ The resulting set of weighted particles is a valid approximation of the measure of $m(\vtheta)=p(\vtheta)-\mu_M(\vtheta)$ and can be used to approximate any expectation under it \cite{fearnhead2010random,tran2013importance,llorente2022optimality}.  Namely, increasing the samples $N$, the  weighted approximate better and better the measure of  $m(\vtheta)$ and, increasing $M$, $m(\vtheta)\rightarrow p(\vtheta)$. 
 This noisy IS algorithm is also called {\it random weight} IS or IS {\it squared} in the literature.
 \newline
  It can be shown that both algorithms, N-MH and N-IS, work on an extended space where $m(\vtheta)=p(\vtheta)-\mu_M(\vtheta)$ is a marginal distribution (see App. App. \ref{app_noisyMH} and \ref{App_noisyIS} for a simple proof). 
}

%
%
%

\begin{algorithm}[!h]
	\centering
	\caption{Noisy Metropolis-Hastings (N-MH) algorithms}
	{\footnotesize
		\begin{tabular}{|p{0.95\columnwidth}|}
			\hline
			\begin{itemize}
				\item[1.] \textbf{Inputs:} Initial state $\vtheta_0$ and realization $\widetilde{p}_M( \vtheta_0)$.
				
				\item[2.]\label{} For $t=1,\dots,T$:
				\begin{enumerate}
					\item[(a)]  Sample $\vtheta_\text{prop} \sim \varphi(\vtheta|\vtheta_{t-1})$ and obtain realization $\widetilde{p}_{\text{now}}=\widetilde{p}_M( \vtheta_\text{prop})$.
					
					\item[(b)] In the {\it pseudo-marginal MH} (PM-MH) set $\widetilde{p}_{\text{bef}}=\widetilde{p}_M( \vtheta_{t-1})$; otherwise, in the {\it  Monte Carlo-within-MH}, obtain a new realization {$\widetilde{p}_M$} at $\vtheta_{t-1}$ and set $\widetilde{p}_{\text{bef}}=\widetilde{p}_M( \vtheta_{t-1})$. 
					\item[(c)] With probability 
					\begin{align}
						\label{eq_MHaccProb}
						\alpha(\vtheta_{t-1},\vtheta_\text{prop}) =\min \left\{1,\frac{\widetilde{p}_{\text{now}} \mbox{ } \varphi(\vtheta_{t-1}|\vtheta_\text{prop})}{\widetilde{p}_{\text{bef}} \mbox{ } \varphi(\vtheta_\text{prop}|\vtheta_{t-1})}\right\},
					\end{align}
					accept $\vtheta_\text{prop}$, i.e., set $\vtheta_t = \vtheta_\text{prop}$.					
					Otherwise, reject $\vtheta_\text{prop}$, i.e., set $\vtheta_t = \vtheta_{t-1}$.
				\end{enumerate}			
				\item[3] \textbf{Outputs:} the chain $\{\vtheta_{t}\}_{t=1}^T$.
			\end{itemize}\\
			\hline
		\end{tabular}	
	}
	\label{tab_pseudoMH}
	\vspace{0.2cm}
\end{algorithm}

\begin{algorithm}[!h]
	\centering
	\caption{Noisy importance sampling (N-IS) algorithm}
	{\footnotesize
		\begin{tabular}{|p{0.95\columnwidth}|}
			\hline
			\begin{itemize}
				\item[1.] \textbf{Inputs:} Proposal distribution  $q(\vtheta)$.
				
				\item[2.]\label{} For $n=1,\dots,N$:
				\begin{enumerate}
					\item[(a)]  Sample $\vtheta_n \sim q(\vtheta)$ and obtain realization $\widetilde{p}_M( \vtheta_n)$.

					\item[(b)] Compute 
					\begin{align}\label{eq_ISweightComp}
						w_n = \frac{\widetilde{p}_M( \vtheta_n)}{q(\vtheta_n)}
					\end{align}
				\end{enumerate}	
			
				\item[3] Compute normalized weights: $\bar{w}_n = \frac{w_n}{\sum_{j=1}^Nw_j},\ j=1,\dots,N$.

				\item[4] \textbf{Outputs:} the weighted samples $\{\vtheta_{n},\bar{w}_n\}_{n=1}^N$.
			\end{itemize}\\
			\hline
		\end{tabular}	
	}
	\label{tab_noisyIS}
	\vspace{0.2cm}
\end{algorithm}

\subsection{Accelerating and denoising by surrogate models}

{
\noindent
In many works, the authors suggest the use of surrogate smoothing models $\widehat{p}(\vtheta)$,  to avoid a new realization of $\widetilde{p}_M( \vtheta)$, and build employing the information of previous realizations of $\widetilde{p}_M$ in different $ \vtheta'$. The accelerated schemes are then obtained replacing $\widetilde{p}_M( \vtheta)$ with $\widehat{p}(\vtheta)$ in the Algorithms \ref{tab_pseudoMH} and \ref{tab_noisyIS} above. 
 The motivation is two-fold: 
 \begin{itemize}
\item {\bf denoising:} reduce the randomness (due to the variance of $\widetilde{p}_M$, without increasing $M$);
\item {\bf accelerating:} reduce the computation cost (evaluation of $\widehat{p}(\vtheta)$ is often cheaper than obtaining a new realization of $\widetilde{p}_M( \vtheta)$).
\end{itemize}
 The surrogate model can be built by a parametric or non-parametric regression classes of methods. We start from the latter, which often provides the better performance.  }
\newline
\newline
{{\bf Non-parametric regression.}} The vanilla schemes described above can be improved by applying non-parametric regression models  $\widehat{p}(\vtheta)$ from the noisy realizations.  More specifically, considering the set of $J$ observed points $\{\vtheta_i, \widetilde{p}_M( \vtheta_i)\}_{i=1}^J$, we apply a regression model for obtaining $\widehat{p}(\vtheta)$, e.g., 
{global models such as radial basis functions  and Gaussian process (GP) regression, or local models such as (weighted) k-nearest neighbors (k-NN), local-GP and local polynomial regression \cite{rasmussen06,bishop,wendland2004scattered,gramacy2015local,kohler2002universal}.  The choice of nonparametric construction allows to improve the surrogate $\widehat{p}(\vtheta)$ as the number of points $J$ grows.
Namely, We assume that with a  non-parametric regression model, we can obtain $\widehat{p}(\vtheta)$ converges to $m(\vtheta)$ as $J\rightarrow \infty$. Increasing also $M$,  we have  $\mu_M(\vtheta)$ and $\widehat{p}(\vtheta)\rightarrow p(\vtheta)$ for all $\vtheta$. } The locations of the nodes $\vtheta_i$ can be chosen appropriately  for ensuring the convergence when $J \to \infty$, under mild conditions {\cite{schaback1995error,stuart2018posterior,pronzato2012design,devroye2017measure}}. Necessary conditions over   $\widehat{p}(\vtheta)$ are shown below.

{\rem 
A necessary condition is that the construction of $\widehat{p}(\vtheta)$ must be strictly positive, $\widehat{p}(\vtheta)>0$,  for all $\vtheta$ where $m(\vtheta)>0$. { we also require that $\int_{\bm \Theta} \widehat{p}(\vtheta) d\vtheta < \infty$.}}
{
\newline
\newline
It is important to remark that, in a non-parametric construction as  $J$ increases, the computational cost of evaluating $\widehat{p}(\vtheta)$ grows as well.  
{\rem 
Evaluating $\widehat{p}(\vtheta)$ in any $\vtheta$ should be cheaper or equal-costly than obtaining new realization $\widetilde{p}_M( \vtheta)$. In the case of ``equal-cost'' the benefit of using $\widehat{p}(\vtheta)$ is only in the reduction of randomness/variance. }
}
\newline
\newline
{
\noindent{\bf Other approximations.} Although our focus is on nonparametric surrogates, other possibilities exist for approximating the evaluation of $p(\vtheta)$.
For instance, one can build a local or global Gaussian approximation to $p(\vtheta)$ via Laplace approximation, expectation-propagation or variational inference \cite{bishop,gomez2018markov}. 
{ However, in this case, we $\widehat{p}(\vtheta)$ generally {\it does not} converge to $m(\vtheta)$ as $J\rightarrow \infty$, but the the computational cost of evaluating $\widehat{p}(\vtheta)$ remains invariant.}
In specific applications,  domain-specific approximations have been designed \cite{christen2005markov,golightly2015delayed,lowe2023accelerating}. More generally, in the MCMC context, approximations of the whole acceptance ratio can be built and used instead of the true one \cite{alquier2016noisy}. 
}
\newline
\newline
{{\bf Selection of the nodes.}} Clearly, the selection of the design nodes $\{\vtheta_i, \widetilde{p}_M( \vtheta_i)\}_{i=1}^J$ is a very important point. 
Several strategies have been proposed for obtaining the set of design nodes are, for instance, running a pilot MCMC run \cite{drovandi2018accelerating}, applying Bayesian experimental design algorithms \cite{jarvenpaa2021parallel,svendsen2020active}, space-filling heuristics \cite{conrad2016accelerating,llorente2020adaptive}, or optimization \cite{bliznyuk2008bayesian,pronzato2012design}. { The concept of an acquisition function is often employed \cite{llorente2021deep,llorente2020adaptive,svendsen2020active} (see for instance App. \ref{EmuLogLikeSect}).  As we will see below,} in the iterative refinement schemes, the path of the chain can also be used  to update the surrogate, either \textit{directly} by including some of the states of the chain \cite{ying2020moving}, or \textit{indirectly} by guiding the search of design points with other techniques \cite{conrad2016accelerating}. 
\newline
\newline
We remark that the use of a surrogate is beneficial  for working with both costly and noisy target pdfs. 
In the following, we review different MCMC and IS approaches that can deal with noisy and expensive target distributions. Some of these methods have been originally proposed only for the expensive or just noisy case (i.e., in a more restricted range of application), but 
they can address the complete problem considered in this work. 
 The main notation of this work is summarized below:

{
\begin{center}
	\begin{tabular}{|c|c|c|} 
		\hline
		\multirow{2}{*}{\textbf{Density of interest}}
		 &  \textbf{Noisy realization/}  & \multirow{2}{*}{\textbf{Surrogate}}    \\
		&  \textbf{Estimation}  &   \\
		\hline
		\hline 
		\multirow{5}{*}{$p(\vtheta)$} 	& \multirow{2}{*}{$\widetilde{p}_M( \vtheta)$} & \multirow{5}{*}{$\widehat{p}(\vtheta)$ built using $\{\vtheta_i, \widetilde{p}_M( \vtheta_i)\}_{i=1}^J$} 		 \\ 
		& & \\
		\cline{2-2}
		& \multirow{2}{*}{$\mathbb{E}[\widetilde{p}_M(\vtheta)] =m(\vtheta) =p(\vtheta)-\mu_M(\vtheta)$}  & \\
			& \multirow{2}{*}{$\texttt{var}[\widetilde{p}_M(\vtheta)] = s_M^2(\vtheta)$}  &  \\ 
		&  & \\
		\hline
	\end{tabular}
\end{center}
}

\section{Overview and generic scheme}\label{GenScheme}

%
%
%
%
%
%
%
%

\begin{figure}[h!]
	\centering
	\includegraphics[width=0.7\textwidth]{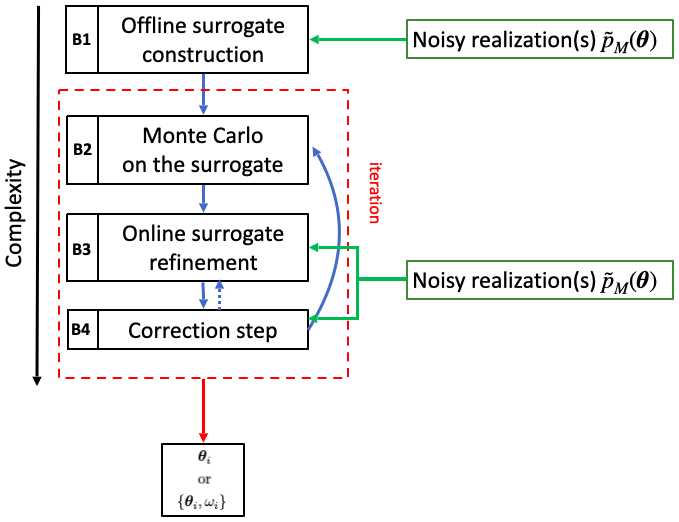}
	\caption{General outline of the schemes considered in the work.} \label{fig_la_gran_polla}
\end{figure}

In this section, we present a general scheme that combines Monte Carlo with the use of surrogates, which encompasses most of the methods proposed in the literature for costly or noisy target pdfs.  Moreover, we distinguish three main classes of methods:
{ We have classified the methods in 3 main families of methods:
\begin{itemize}
\item   \textbf{(C1)} Two-stage schemes,
\item \textbf{(C2)} iterative refinement schemes, and
\item \textbf{(C3)} exact schemes.
\end{itemize} }
\noindent
In this section, we provide a brief description of each of them.
\newline
A graphical representation of a generic scheme is given in Figure \ref{fig_la_gran_polla}, that is composed of a series of four blocks.
Each approach in the literature is formed by a different combination of blocks { as shown in   Table \ref{tab_bloques}, that gives a summary of the relationship between the three families and all the blocks given in Figure \ref{fig_la_gran_polla}.} The three classes C1, C2, C3 have in common the block B2, i.e., performing one or more Monte Carlo iterations (e.g., MH or IS) with respect to (w.r.t.) the surrogate $\widehat{p}(\vtheta)$ instead of using the realizations $\widetilde{p}_M( \vtheta)$. 

{\rem \label{rem_nonparam}
Note that block B2 can be viewed as sampling from a non-parametric proposal density. Furthermore, the application of Monte Carlo in block B2 could be substituted with a direct sampling of the surrogate when it is possible \cite{martino2018adaptive}.
}
\newline
\newline
Blocks B1 and B3 refer to the two possible strategies {(offline or iteratively within the Monte Carlo steps)} for building the surrogate. The former considers an offline construction, that is totally independent of the Monte Carlo algorithm that will be run afterwards. The latter construction aims to build the surrogate online, i.e., during the Monte Carlo iterations. Lastly, Block B4 refers to making a correction for the fact that we are  working w.r.t.\ $\widehat{p}(\vtheta)$, and ultimately implies obtaining a noisy realization $\widetilde{p}_M( \vtheta)$. 
In the rest of the section, we present each of these families in increasing order of complexity.

\begin{table}[h!]
{
	\centering
	\caption{{\small { Relationship between the four main classes and the blocks (B1, B2, B3 and B4) enumerated in Figure \ref{fig_la_gran_polla}. In parenthesis, we point out the blocks that are optional to each family of methods.}}}  \label{tab_bloques}
	\vspace{-0.2cm}
	\begin{tabular}{|l|| c | c | c|} 
		\hline
		\textbf{Blocks} & {\bf Two-stage} &  {\bf  Iterative refinement} &  {\bf  Exact w.r.t.\ $p(\vtheta)-\mu_M(\vtheta)$}  \\ [0.5ex] 
		\hline
		{\bf B1:} Offline surr. construction	& \checkmark  & (\checkmark)  &  (\checkmark)    \\ [1ex] 
		{\bf B2:} MC on surrogate	& \checkmark  & \checkmark  & \checkmark  \\ [1ex] 
		{\bf B3:} Online surr. refinement	&  &  \checkmark &  (\checkmark)  \\ [1ex] 
		{\bf B4:} Correction step	&  &  & \checkmark  \\ [1ex] 
		\hline
	\end{tabular}
	}
\end{table}

{
\subsection{Two-stage schemes}
}


This scheme includes blocks B1 and B2 in Figure \ref{fig_la_gran_polla}.
A \textit{two-stage}  scheme consists in running Monte Carlo algorithm on a fixed surrogate, that has been built offline, i.e., before the start of the algorithm.
This scheme is preferred when the computational budget is limited in advance, so it is all devoted to the surrogate construction. 
This scheme is very common in, e.g., the calibration of expensive computer codes \cite{bliznyuk2008bayesian,cleary2021calibrate}. 
The estimators derived from this scheme are biased, i.e., {$\mu_M(\vtheta)\neq 0$}. However, since this scheme does not imply obtaining costly realizations $\widetilde{p}_M(\vtheta)$ in the second stage, the algorithms can be run for many iterations and produce estimators with low variance. 
For this scheme to be worth, the decrease in variance must compensate the presence of bias.
Recent methods proposed in the literature follow this scheme \cite{drovandi2018accelerating,park2020function,jarvenpaa2021parallel,jarvenpaa2019abc,Gutmann15}.
\newline
{ Regarding the  offline construction of the surrogate, different strategies can be used  to select the design points: for instance, a space-filling design \cite{pronzato2012design}, or pilot runs of MCMC steps \cite{drovandi2018accelerating,park2020function}, or using an optimization algorithm to locate high-probability regions  \cite{bliznyuk2008bayesian}. Furthermore, active learning schemes can be also employed \cite{jarvenpaa2019abc,jarvenpaa2021parallel,Gutmann15,jarvenpaa2021parallel,kandasamy2017query,wang2018adaptive}.}


{
\subsection{Iterative refinement schemes}
}


\noindent
This second scheme comprises blocks B1 (optionally), B2 and B3  in Figure \ref{fig_la_gran_polla}.
It considers iteratively building the surrogate along with the execution of the Monte Carlo algorithm, i.e., $\widehat{p}_t$ depends on $t$.  In every iteration, a test is performed in order to decide if we update the surrogate (i.e., obtain a new noisy realization $\widetilde{p}_M( \vtheta)$). The surrogate refinement can be made at the end and/or beginning of the iteration (i.e., block 3 could be placed before and/or after block B2).
An initial surrogate $\widehat{p}_{0}(\vtheta)$ could be built offline by using some of the strategies of the methods from the previous scheme. 
\newline
{
The use of online surrogates within this class/family of Monte Carlo algorithms has two main motivations.  
Firstly, an improvement of the surrogate over time reduces the discrepancy between  $\widehat{p}_t(\vtheta)$ and $m(\vtheta)=\widehat{p}(\vtheta)-\mu_M(\vtheta)$.}
Generally speaking, if the surrogate is improved infinitely often and in a suitable way (e.g., with a space-filling strategy), the error between the surrogate $\widehat{p}_{t}(\vtheta)$  and $m(\vtheta)$ will approach zero, i.e., is asymptotically exact \cite{conrad2016accelerating,conrad2018parallel,ying2020moving}. 
Clearly, constantly changing the target density within a Monte Carlo algorithm difficult its theoretical analysis. Moreover, in MCMC algorithms, updating the surrogate using past states of the chain produces the loss of Markov property, so (as in the adaptive MCMC literature) one needs to carefully address this point \cite{ying2020moving,conrad2016accelerating}. 
{Secondly, the surrogate construction is driven by the Monte Carlo algorithm, namely, the surrogate is refined in regions  discovered by the algorithm along the iterations. 
This also represents an opportunity to improve the accuracy of the surrogate and reduce the bias by actively selecting points in the neighborhood of the current samples. 
Local surrogate constructions fit very well in this scenario \cite{conrad2016accelerating,davis2022rate}. 
}
\newline
\newline
Some proposed methods that follow this scheme are \cite{conrad2016accelerating,conrad2018parallel,davis2022rate,ying2020moving,jarvenpaa2021approximate}.
In \cite{conrad2016accelerating,conrad2018parallel}, a local GP or polynomial approximation is built on $\log p(\vtheta)$ and refined over the MCMC iterations by using space-filling heuristics. 
{
In \cite{davis2022rate}, the authors extend the work of \cite{conrad2016accelerating,conrad2018parallel} and study optimal online refining strategies of local polynomial 
approximations within MCMC, that balance the decay in surrogate bias and Monte Carlo variance.}  In \cite{jarvenpaa2021approximate}, a GP regression model of $\log p(\vtheta)$ is built online by maximizing acquisition functions derived using Bayesian experimental design in order to decrease the uncertainty in the computation of the MH accept-reject test. This algorithm could be considered as a two-stage procedure (previously described) if we use a pilot run for the construction of the surrogate.
{ In \cite{ying2020moving}, a nearest neighbor approximation of $p(\vtheta)$ is built online in MCMC by including the points that are accepted at each iteration.}

{
\subsection{Exact schemes}
}


\noindent
This scheme includes  blocks (optionally) B1, B2, (optionally) B3  and B4 in Figure \ref{fig_la_gran_polla}.
The main difference w.r.t.\ the previous schemes is the correction step. At some iterations of the method, we obtain a noisy realization $\widetilde{p}_M(\vtheta)$, in order to ensure the correctness of the algorithm, which will approximate and/or converge to $m(\vtheta)=p(\vtheta)-\mu_M(\vtheta)$. Since working with $\widehat{p}(\vtheta)$  is cheaper than obtaining new realizations $\widetilde{p}_M( \vtheta)$, this scheme reduces the computation time. 
However,  the fact that a new realization $\widetilde{p}_M( \vtheta)$ has to be obtained for every ``correction'' usually prevents significant computational savings.
This scheme can be used with a fixed offline-built surrogate, an online surrogate or combination of both.
{
The correction step in MCMC schemes (specifically, the MH algorithm) consists in introducing a second accept-reject test  where $\widehat{p}(\vtheta)$ now plays the role of proposal density (see Algorithm \ref{tab_FoxLiu}). 
}These two-step MH algorithms are known as {\it MH with delayed acceptance} (DA) \cite{christen2005markov,banterle2019accelerating}; see next section for a detailed description.
{
In IS, a set of samples distributed from $\widehat{p}(\vtheta)$ can be corrected by assigning another weights proportional to the ratio $\frac{\widetilde{p}_M( \vtheta)}{\widehat{p}(\vtheta)}$.}
{Hence, in all the correction factors the surrogate $\widehat{p}(\vtheta)$ is considered as a (non-parametric) proposal density} within a Monte Carlo method with target density $m(\vtheta)$. 
\newline
{\rem
	Note that the online improvement of the surrogate corresponds to the adaptation of the equivalent proposal of block B2 (see Remark \ref{rem_nonparam}) using not only the information of past samples, but also the history of noisy evaluations of the target.	 
}
\newline
\newline
Some examples  of methods leveraging surrogate models to produce efficient proposals in the literature are the following (mostly in the non-noisy - but costly - context, i.e., $\widetilde{p}_M( \vtheta)=m(\vtheta)=p(\vtheta)$).
{
Works that employ surrogates within DA in the noisy context are \cite{sherlock2017adaptive,quiroz2018speeding}.
}
DA schemes rely on approximate sampling from the surrogate via one MH step. 
Other works consider an standard MH algorithm where the surrogate is sampled with direct methods   \cite{martino2018adaptive}. A rejection sampling (RS) scheme for sampling the surrogate is applied in \cite{zhang2019accelerating}, where a kriging-based surrogate is built within a delayed rejection MH \cite{haario2006dram}.
In the IS context, the authors in \cite{llorente2021deep} propose sampling the surrogate with IS resampling steps, and then weigh the resulting samples w.r.t.\ the true target. 
\newline
\newline
Table \ref{tab_exactos} provides some examples of methods belonging to ``exact'' family, specifying the type of Monte Carlo technique employed in the blocks B2 and B4 in Figure \ref{fig_la_gran_polla}.Finally, Table \ref{fig_class_methods} shows several works classified into the three presented classes.

\begin{table}[h!]
	\centering
	\caption{	
	Summary of specific examples of ``exact'' algorithms, attending to the Blocks B2 and B4 in Figure \ref{fig_la_gran_polla}.  } \label{tab_exactos}
	\vspace{-0.2cm}
	\begin{tabular}{| c || c | c |} 
		\hline
		\textbf{Exact algorithms} & Block B2 & Block B4 \\ [0.5ex] 
		\hline
		Sticky MCMC \cite{martino2018adaptive} & direct & MCMC \\ [1ex] 
		Noisy Deep IS \cite{llorente2021deep} & IS & IS\\ [1ex] 
		Kriging AIS \cite{balesdent2013kriging}& MCMC & IS\\ [1ex] 
		Delayed-acceptance MH \cite{christen2005markov,banterle2019accelerating}& MCMC & MCMC \\ [1ex] 
		Kriging-based delayed rejection MH \cite{zhang2019accelerating}& RS & MCMC \\ [1ex]
		\hline
	\end{tabular}
	\end{table}


\begin{table}
\caption{Several works in the literature classified into the three presented classes.}\label{fig_class_methods}
\vspace{-0.3cm}
	\begin{center}
		{
		\begin{tabular}{| c | c | c|} 
			\hline
			{\bf Two-stage} & {\bf Iterative refinement} & {\bf Exact} \\ [0.5ex] 
			\hline\hline
			\citet{bliznyuk2008bayesian} & \citet{conrad2016accelerating} & \citet{christen2005markov} \\ 
			\citet{jarvenpaa2021parallel} & \citet{ying2020moving}  & \citet{sherlock2017adaptive} \\
			\citet{drovandi2018accelerating} & \citet{jarvenpaa2021approximate} & \citet{llorente2021deep} \\
			\citet{wang2018adaptive}   & \citet{davis2022rate} & \citet{martino2018adaptive} \\
			\hline
		\end{tabular}

		}	\end{center}
	
\end{table}

\noindent
\textbf{Honorable mentions.}  Other ways of using surrogates to improve Monte Carlo methods that do not compromise the exactness are, e.g., HMC with gradient computations based on the surrogate \cite{rasmussen2003gaussian}. 
In \cite{fielding2011efficient}, the authors introduce extensions of the previous idea to multimodal scenarios by combining it with parallel tempering, where only the lowest temperature chain addresses the true posterior while the other chains at higher temperatures work with surrogates.

{
{\rem The IS-based algorithms return weighted samples. These weighted samples can be directly used for approximating posterior expectations as in a quadrature  scheme \cite{elvira2020importance,llorente2020adaptive}. Otherwise, they can be converted into unweighted samples by resampling them \cite{luengo2020survey}.
}
}

{
\subsection{Dependency on the surrogate in the three classes}


A preliminary necessary observation is that the type of surrogate should selected as in a regression problem: namely, considering the a-priori information regarding smoothness and other features of the posterior to emulate. 
\newline
In the two-stage scheme, the process of building the surrogate and performing the sampling is separated. During the initial stage, the primary objective is to minimize the bias of the surrogate. In the subsequent stage, our focus shifts to reducing the variance of the Monte Carlo approximation of the surrogate. Due to the inexpensive nature of evaluating the surrogate, it becomes possible to obtain Monte Carlo approximations with extremely low variance. However, employing surrogates introduces a bias into the final estimators.
\newline
In the iterative refinement scheme, surrogates are constructed while executing the Monte Carlo algorithms. Therefore, our aim is to minimize both bias and variance simultaneously. 
It is crucial to carefully consider the impact of modifying the surrogate throughout the iterations on the convergence of the algorithms, since constantly altering the target distribution of the algorithms can potentially difficult their ability to converge successfully.  In the exact scheme, the surrogate serves as a proposal, eliminating any bias, and allowing the algorithms to directly target the distribution of interest. However, it is important to note that a poor construction of the surrogate can hinder the exploration process by Monte Carlo algorithms, and in some cases, even jeopardize their convergence.
\newline
Last but not least, avoiding the overfitting in the construction of the surrogate is particularly beneficial  in the initial steps of the all classes of algorithms, fostering the exploration of the space. In non-parametric regression can be done easily controlling some parameter: the power of noise in GP regression, as an example, or the number of neighbors in k-nearest neighbor  (kNN) regression. For instance, in a kNN approximation, one could start with a bigger number of neighbors  and decreases this number as the iterations grow.
}

\section{Specific instances of noisy Monte Carlo methods}\label{SectionSpecificExample}

In this section, we describe in details some specific techniques which are included in the generic scheme described in the previous section.
They are Monte Carlo algorithms that were introduced mainly in the context  of costly, but non-noisy, targets, but their extension to the noisy setting is straightforward. 
We focus on the iterative and exact families of methods, but it should be noted that the strategies for building offline surrogates (from the algorithms within the two-stage scheme) could also be used to initialize the surrogates and hence further improve these algorithms. 
\newline
\newline
\textbf{MH schemes on iterative surrogate.} 
A generic MH algorithm targeting a surrogate that is refined over $T$ iterations is given in Algorithm \ref{tab_LucaMoving}.  This algorithm falls within the iterative refinement scheme from the previous section.
We also include  in Algorithm \ref{tab_LucaMoving} different variants. Indeed,
at each iteration, the surrogate is updated with probability $\rho_\text{update}$,  obtaining a noisy realization and including it in the set of active nodes. The different variants are obtained by designing a different probability $\rho_\text{update}$ and deciding the search strategy for finding a suitable new node $\vtheta^*$.
\newline
Note that the updating probability could depend on many features, e.g., on  the current surrogate $\rho_\text{update}=\rho^{(t)}_\text{update}(\widehat{p}_{t-1},\psi)$ and other parameters $\psi$.
{  For instance, the surrogate can be updated every time the estimated error of the surrogate or the acceptance probability is above some threshold \cite{conrad2016accelerating,jarvenpaa2021approximate}}.
The new point to be included, $\vtheta^*$, can be chosen by different strategies.
{  For instance, such that the empty space within the neighborhood of current state or the expected uncertainty of $\alpha_\text{MH}$ is reduced the most \cite{conrad2016accelerating,davis2022rate,jarvenpaa2021approximate}.}
\newline
As an example, in this work we will specifically consider and compare two basic algorithms. The first one corresponds to $\rho_\text{update}=1$, while the second one  considers $\rho_\text{update}=\alpha^{(t)}_\text{MH}$ \cite{ying2020moving}. 
In both, we consider the simple choice $\vtheta^*=\vtheta_{\texttt{prop}}$, i.e., the new node is the proposed state at that iteration. 
The updating block could be also placed before the MH acceptance test and repeated until some criterion is met \cite{conrad2016accelerating,jarvenpaa2021approximate}. 

\begin{algorithm}[!h]
	\centering
	\caption{Metropolis-Hastings on surrogate with iterative refinement (MH-S)}
	{\footnotesize
		\begin{tabular}{|p{0.95\columnwidth}|}
			\hline
			\begin{itemize}
				\item[1.] \textbf{Inputs:} Initial state $\vtheta_0$ and initial surrogate $\widehat{p}_{0}(\vtheta)=\widehat{p}_{0}(\vtheta; \mathcal{S}^{(0)})$.
				
				\item[2.]\label{} For $t=1,\dots,T$:
				\begin{enumerate}

					\item[(a)]  Sample $\vtheta_\text{prop} \sim \varphi(\vtheta|\vtheta_{t-1})$.
					
					\item[(b)] With probability 
					\begin{align}
						\label{}
						\alpha(\vtheta_{t-1},\vtheta_\text{prop}) =\min \left\{1,\frac{\widehat{p}_{t-1}(\vtheta_\text{prop}) \varphi(\vtheta_{t-1}|\vtheta_\text{prop})}{\widehat{p}_{t-1}(\vtheta_{t-1}) \varphi(\vtheta_\text{prop}|\vtheta_{t-1})}\right\},
					\end{align}
					accept $\vtheta_\text{prop}$, i.e., set $\vtheta_t = \vtheta_\text{prop}$.
					Otherwise, reject $\vtheta_\text{prop}$, i.e., set $\vtheta_t = \vtheta_{t-1}$.
						
					\item[(c)] With probability $\rho_\text{update}$: 	 {(otherwise, with probability $1-\rho_\text{update}$, skip to the next iterations)}						
					\begin{enumerate}
								\item {Select (according to some strategy)} $\vtheta^\star$ and obtain the realization $\widetilde{p}_M( \vtheta^\star)$.
								\item Update design nodes set $\mathcal{S}^{(t)} = \mathcal{S}^{(t-1)} \cup \{\vtheta^\star,\widetilde{p}_M( \vtheta^\star)\}$.
							\end{enumerate}
				{\item[(d)] Build the surrogate $\widehat{p}_{t}(\vtheta)=\widehat{p}_{t}(\vtheta;\mathcal{S}^{(t)})$ from $\mathcal{S}^{(t)}$.}				
				\end{enumerate}
							
				\item[3] \textbf{Outputs:} The chain $\{\vtheta_{t}\}_{t=1}^T$ and the final surrogate $\widehat{p}_{T}(\vtheta)$.
			\end{itemize}\\
			\hline
		\end{tabular}	
	}
	\label{tab_LucaMoving}
	\vspace{0.2cm}
\end{algorithm}

\vspace{0.2cm}
\noindent\textbf{Delayed-acceptance Metropolis-Hastings.}
The DA-MH algorithm is a modified MH algorithm (also called {\it `two-step MH'} or {\it`MH with early rejection'}  \cite{christen2005markov,banterle2019accelerating}) where, at each iteration, the proposed state $\vtheta_\text{prop}$ is undergone to two MH accept-reject tests.
We consider here delayed-acceptance  pseudo-marginal MH (DA-PM-MH), where noisy evaluations are recycled.
At each iteration, the proposed state is tested first against $\widehat{p}(\vtheta)$ (i.e., block B2 is a MH step on the surrogate) and, upon acceptance, then against $\widetilde{p}_M( \vtheta)$ (i.e., block B4 is a noisy MH step). 
\newline
\newline
The computational savings occur when $\vtheta_\text{prop}$ is rejected in the first test, since it avoids performing the second MH test and computing the costly noisy realization $\widetilde{p}_M( \vtheta_\text{prop})$.
In this work, we consider also a general version DA-PM-MH (also called {surrogate transition method} \cite{liu2008monte}) that allows for multiple iterations w.r.t.\ $\widehat{p}$ in the first step.
The details are given in Algorithm \ref{tab_FoxLiu}. The standard DA-PM-MH algorithm is recovered setting $T_\text{surr}=1$.
The standard DA-PM-MH has always a lower acceptance than vanilla PM-MH \cite{banterle2019accelerating}, but can provide better performance.
However, for $T_\text{surr}\ge 1$, the acceptance probability can be higher than in the standard MH.
The first step aims at obtaining a good candidate $\vxi_{T_\text{surr}}$ by sampling (via MCMC) from a proposal density $\widehat{p}$ built by a (usually non-parametric) surrogate model. The candidate sample  $\vxi_{T_\text{surr}}$   is then employed in a MH test w.r.t.\ $\widetilde{p}$. We can interpret the DA-MH as a two-step algorithm where, in the first step, samples approximately distributed as the surrogate are generated.
Thus, other algorithms such as sticky MCMC \cite{martino2018adaptive} can be considered as `ideal' version of DA-MH, since the samples are drawn directly from the surrogate (i.e. the acceptance probability in the first step is always one).


\begin{algorithm}[!h]
	\centering
	\caption{DA-PM-MH algorithm}
	{\footnotesize
		\begin{tabular}{|p{0.95\columnwidth}|}
			\hline
			\begin{itemize}
				\item[1.] \textbf{Inputs:} Initial state $\vtheta_0$, initial realization $\widetilde{p}_M( \vtheta_0)$, surrogate $\widehat{p}_0(\vtheta;\mathcal{S}^{(0)})$, and number of `inner' iterations $T_\text{surr}$.
				\item[2.] For $t=1,\dots,T$:
				\begin{itemize}
			
					\item[(a)]  Starting from $\vtheta_{t-1}$, run $T_\text{surr}$ iterations of MH with respect to $\widehat{p}_{t-1}(\vtheta)$. That is, set $\vxi_0=\vtheta_{t-1}$ and do for $k=1,\dots,T_\text{surr}$:
					\begin{itemize}
						\item[(i)] Sample $\vxi' \sim q(\vtheta|\vxi_{k-1})$
						\item[(ii)] With probability
						$$
						\alpha_1(\vxi_{k-1},\vxi')=\min \left\{1,\frac{\widehat{p}_{t-1}(\vxi') \varphi(\vxi_{k-1}|\vxi')}{\widehat{p}_{t-1}(\vxi_{k-1}) \varphi(\vxi'|\vxi_{k-1})}\right\},
						$$
						accept $\vxi'$, i.e., set $\vxi_k = \vxi'$. Otherwise, reject $\vxi'$, i.e., set $\vxi_k=\vxi_{k-1}$.
					\end{itemize}

					\item[(b)] Set $\vtheta_\text{prop}=\vxi_{T_\text{surr}}$, and accept it with probability
					$$
						\alpha_1(\vtheta_{k-1},\vtheta_\text{prop})=\min \left\{1,\frac{\widetilde{p}_M( \vtheta_\text{prop})
						\widehat{p}_{t-1}(\vtheta_{t-1})
					}{
						\widetilde{p}_M( \vtheta_{t-1})
						\widehat{p}_{t-1}(\vtheta_\text{prop})
					}\right\},
					$$
					i.e., set $\vtheta_t = \vtheta_\text{prop}$. Otherwise, reject $\vtheta_\text{prop}$, i.e., set $\vtheta_t = \vtheta_{t-1}$. 

					\item[(c)] With probability $\rho_\text{update}$: {(otherwise, with probability $1-\rho_\text{update}$, skip to the next iterations)}
					\begin{enumerate}
						\item {Select (according to some strategy)} $\vtheta^\star$ and obtain the realization $\widetilde{p}_M( \vtheta^\star)$.
						\item Update design nodes set $\mathcal{S}^{(t)} = \mathcal{S}^{(t-1)} \cup \{\vtheta^\star,\widetilde{p}_M( \vtheta^\star)\}$.
					\end{enumerate}
	{\item[(d)] Build the surrogate $\widehat{p}_{t}(\vtheta)=\widehat{p}_{t}(\vtheta;\mathcal{S}^{(t)})$ from $\mathcal{S}^{(t)}$.}			
				\end{itemize}
				
				\item[3] \textbf{Outputs:} The chain $\{\vtheta_t\}_{t=1}^T$.			
			\end{itemize}\\
			\hline
		\end{tabular}	
	}
	\label{tab_FoxLiu}
	\vspace{0.2cm}
\end{algorithm}

\vspace{0.2cm}
\noindent\textbf{Noisy Deep Importance Sampling (N-DIS).}
The Deep Importance Sampling (DIS) is an adaptive IS scheme introduced in \cite{llorente2021deep}, which uses a non-parametric surrogate as its proposal density. 
It can be seen as a multivariate extension of the technique in \cite{martino2018adaptive}.
Here, we consider a noisy version of DIS, which is described in Algorithm \ref{tab_RADIS}. Again the underlying idea is to use the surrogate $\widehat{p}(\vtheta)$ as proposal density. 
For sampling from $\widehat{p}(\vtheta)$, N-DIS employs a Sampling Importance Resampling (SIR)  approach \cite{Rubin88}, using an auxiliary/parametric proposal, $q(\vtheta)$ (i.e., block B1 is a SIR scheme). More specifically,  a set  $\{{\bf y}\}_{\ell=1}^L$ is sampled from  $q(\vtheta)$, with $L\gg1$,  and weighted according to $\widehat{p}$. Then, $N$  resampling steps ($N\ll L$) are performed to obtain $\{\vtheta_i\}_{i=1}^N$, that are approximately distributed as $\widehat{p}(\vtheta)$ \cite{Rubin88}. 
These samples are finally weighted considering the corresponding realizations $\widetilde{p}_M( \vtheta_i)$ (i.e., block B4 is a IS iteration). Thus, N-DIS is as a two-stage IS scheme, where the inner IS stage is employed to draw from the surrogate $\widehat{p}$. 
Furthermore, N-DIS is an iterative algorithm where the previous steps are repeated and the set $\{\vtheta_i\}_{i=1}^N$ is used to refine the surrogate at each iteration. 
Hence, compared to a standard IS scheme, N-DIS improves the performance by using a non-parametric surrogate proposal density $\widehat{p}(\vtheta)$ that gets closer and closer to $m(\vtheta)$.
Moreover, N-DIS could be interpreted as an IS version equivalent to the DA-MH algorithm. Note that, N-DIS uses deterministic mixture IS weights in Eq. \eqref{MIS_eq} which provide more stability in the results \cite{cornuet2012adaptive}.


\begin{algorithm}[!h]
	\centering
	\caption{N-DIS algorithm (DIS with noisy realizations)}\label{tab_RADIS}
	{\footnotesize
		\begin{tabular}{|p{0.95\columnwidth}|}
			\hline
			\begin{itemize}
				\item[1.] \textbf{Inputs:} Proposal distribution  $q(\vtheta)$ and initial surrogate $\widehat{p}_{0}(\vtheta)=\widehat{p}_{0}(\vtheta;\mathcal{S}^{(0)})$.
				
				\item[2.]\label{} For $t=1,\dots,T$:
				\begin{enumerate}
					\item[(a)] Sample $\vxi_{t,\ell}\sim q(\vtheta)$, $\ell=1,\dots,L$.

					\item[(b)] Compute $\gamma_{t,\ell} = \dfrac{\widehat{p}_{t-1}(\vxi_{t,\ell})}{q(\vxi_{t,\ell})}$, $\ell=1,\dots,L$.
					
					\item[(c)] Resample $\vtheta_{t,n} \sim \{\vxi_{t,\ell}\}_{\ell=1}^L$, $n=1,\dots,N$, with probabilities proportional to $\{\gamma_{t,\ell}\}_{\ell=1}^L$ (with $N \ll L$).
					
					\item[(d)] Obtain noisy realizations and compute ($n=1,\dots,N$),
					\begin{equation}\label{MIS_eq}
						w_{t,n}= \dfrac{\widetilde{p}_M( \vtheta_{t,n})}{\frac{1}{t} \sum\limits_{\tau=0}^{t-1}\widehat{p}_{\tau}(\vtheta_{t,n})}.
					\end{equation}	
					
					\item[(e)] Update design nodes set $\mathcal{S}^{(t)} = \mathcal{S}^{(t-1)}\cup \{(\vtheta_{t,n},\widetilde{p}_M( \vtheta_{t,n}))\}_{n=1}^N$.
					
			{\item[(f)] Build the surrogate $\widehat{p}_{t}(\vtheta)=\widehat{p}_{t}(\vtheta;\mathcal{S}^{(t)})$ from $\mathcal{S}^{(t)}$.}
						
				\end{enumerate}	
				
				\item[3] Compute normalized weights: {$\bar{w}_{t,n} = \frac{w_{t,n}}{\sum_{\tau=1}^T\sum_{j=1}^Nw_{j,\tau}}, \ n=1,\dots,N, \ t = 1,\dots, T$}.

				\item[4] \textbf{Outputs:} the weighted samples $\{\vtheta_{t,n},\bar{w}_{t,n}\}_{n=1}^N$ {for t=1,\dots,T} and the final surrogate $\widehat{p}_{T}(\vtheta)$.
			\end{itemize}\\
			\hline
		\end{tabular}	
	}
	\vspace{0.2cm}
\end{algorithm}

\newpage
\newpage

%

%

{
\section{Trade-offs and other theoretical analyses}\label{TradeoffSect}

Clearly, the overall computational cost and speed of the algorithms described above depends on the following main factors:
\begin{itemize}
\item number of iterations ($T$) of MCMC steps and or IS samples ($N$);
\item number of samples ($M$) used for obtaining $\widetilde{p}_M(\vtheta)$;
\item number of nodes ($J$) employed for building the surrogate $\widehat{p}(\vtheta)$.
\end{itemize}
Hence, we have different trade-offs for computational cost. For instance, we can reduce the bias $\mu_M(\vtheta)$ and variance $s_M^2(\vtheta)$ of $\widetilde{p}_M(\vtheta)$ increasing $M$, or use a long chain increasing $T$ considering acceptable a certain level of bias and variance of $\widetilde{p}_M(\vtheta)$. This will be clear in the applications described in Sections \ref{sec_abc} and \ref{RLsect}.
\newline
More generally, the efficiency losses of these algorithms with respect to their non-noisy versions have been studied in the literature \cite{doucet2015efficient,sherlock2015efficiency,llorente2022optimality,tran2013importance,fearnhead2010random}. 
Generally, their asymptotic variance is at least as large as their non-noisy versions \cite{andrieu2015convergence,tran2013importance}.
For a fixed computational cost, one can decide to work with unbiased estimates $\widetilde{p}_M(\vtheta)$ with lower variance or running the algorithms for more iterations. In pseudo-marginal MCMC, under the assumptions that the noise of $\log\widetilde{p}(\vtheta)$ is additive Gaussian and independent of $\vtheta$, and the computing time is inversely proportional to the variance, one should work with log-estimates that have a variance of around $1.2$ \cite{doucet2015efficient}.  
Under similar assumptions in noisy IS, the authors in \cite{tran2013importance} derive a closed-form expression for the optimal variance of $\log\widetilde{p}(\vtheta)$ and propose to set the number of particles to match this value. Conversely, for a given noise variance function, noisy IS optimal proposals can be derived \cite{llorente2022optimality}. 
\newline
\newline
In the noiseless setting, the authors in \cite{banterle2019accelerating} study the performance gains of DA schemes, where, more generally, the acceptance probability is a factorized product of more than two terms (this would correspond to, e.g.,  using a {\it cascade} of surrogates). For these algorithms to have good theoretical properties, each factor should be bounded away from zero when the acceptance probability of the standard MH is 1.
To ensure the fulfillment of this condition, they propose a modification of any given factorization. Additionally, they emphasized the benefits of using flattened versions of the $p(\vtheta)$ as surrogates. 
The studies of \cite{sherlock2022variance} also support that for these DA schemes to inherit good properties, the surrogates should generally have heavier tails than the target density. 
In the case where the acceptance probability is factorized into the product of two terms, the work \cite{banterle2019accelerating} shows that the optimal acceptance rate of DA schemes becomes smaller as the cost of computing the surrogate decreases. Conversely, the optimal scale of a random-walk proposal becomes larger when the cost of the surrogate decreases. In other words, when a cheap surrogate is available, it seems beneficial to attempt large jumps from the current state.
\newline
The work of \cite{sherlock2021efficiency} study theoretically DA schemes in the context of unbiased estimations of $p(\vtheta)$ (DA-PM-MH) with random-walk proposals. They confirm the intuition that, provided that the surrogate is inexpensive and reasonably accurate, random-walk DA-PM-MH should be optimally efficient when the scale of the proposal is much larger than that of the equivalent random-walk PM-MH, with larger proposed jumps compared to the surrogate transition algorithm discussed above. Additionally, the authors in \cite{sherlock2021efficiency}  propose a three-step procedure to determine the optimal parameters of random-walk DA-PM-MH. These parameters include the scale of the proposal and the variance of $\widetilde{p}_M( \vtheta)$ (i.e. the number of auxiliary samples for obtaining the unbiased estimation). The procedure involves (1) tuning the scale and the variance of an approximately optimal random-walk  PM-MH, (2) running a random-walk DA-PM-MH using these parameters and computing the average conditional acceptance probability of the second reject test, and (3) looking up a table the corresponding optimal values for the scale and variance.

}

\section{Two specific relevant applications} 
\subsection{Likelihood-free context}  \label{sec_abc}
The Likelihood-free framework in Bayesian inference presents some peculiarities which deserve a specific discussion. We start with a brief description of a generalized approximate Bayesian computation (ABC) scheme in the same fashion of \cite{wilkinson2014accelerating,prangle2016lazy}. 
Given some vector of data ${\bf y}_{\texttt{true}}\in \mathbb{R}^{D_Y}$, in several applications, sampling from a posterior distribution $p(\vtheta)=p(\vtheta|{\bf y}_{\texttt{true}}) \propto \ell({\bf y}_{\texttt{true}}|\vtheta) g(\vtheta)$ is required, where $\ell({\bf y}_{\texttt{true}}|\vtheta)$ represents a likelihood function and $g(\vtheta)$ a prior density. In some context, the pointwise evaluation of    $\ell({\bf y}_{\texttt{true}}|\vtheta)$  is not possible, but we can generate artificial data, ${\bf y}'\sim \ell({\bf y}|\vtheta)$. Hence, we could draw samples in an extended space, $[\vtheta',{\bf y}']$, from the joint pdf $q(\vtheta,{\bf y})=\ell({\bf y}|\vtheta) g(\vtheta)$, drawing first $\vtheta'\sim g(\vtheta)$ and then ${\bf y}' \sim \ell({\bf y}|\vtheta)$.
\newline
The idea behind several ABC algorithms is the following.
  Let us consider the  following extended target pdf in the extended space $[\vtheta,{\bf y}]$,
$$
p_e(\vtheta,{\bf y}|{\bf y}_{\texttt{true}},\epsilon)\propto  h({\bf y}_{\texttt{true}}|{\bf y},\vtheta,\epsilon) \ell({\bf y}|\vtheta) g(\vtheta),
$$
where $h({\bf y}_{\texttt{true}}|{\bf y},\vtheta,\epsilon)\geq 0$ is a {\it surrogate extended likelihood} and $\epsilon>0$ is a positive parameter, chosen by the user.   
In many ABC approaches, different authors consider a simplified version where
$$
h({\bf y}_{\texttt{true}}|{\bf y},\vtheta,\epsilon)=h({\bf y}_{\texttt{true}}|{\bf y},\epsilon),
$$
for instance, $h({\bf y}_{\texttt{true}}|{\bf y},\epsilon)\propto \exp\left(-\frac{||{\bf y}_{\texttt{true}}-{\bf y} ||^2}{2\epsilon^2}\right)$. Hence, we can simplify the previous expression as
$p_e(\vtheta,{\bf y}|{\bf y}_{\texttt{true}})\propto  h({\bf y}_{\texttt{true}}|{\bf y},\epsilon) \ell({\bf y}|\vtheta) g(\vtheta)$. The simplest choice, as in the rejection-ABC scheme, is 
\begin{gather}
\left\{
\begin{split}
&h({\bf y}_{\texttt{true}}|{\bf y},\epsilon)\propto 1 \quad \mbox{ if }  \mbox{  } ||{\bf y}_{\texttt{true}}-{\bf y}||<\epsilon, \\
 &h({\bf y}_{\texttt{true}}|{\bf y},\epsilon)=0 \quad \mbox{ if }\mbox{  }  ||{\bf y}_{\texttt{true}}-{\bf y}||\geq \epsilon.
 \end{split}
 \right.
\end{gather}
Therefore, the ABC target density is
\begin{eqnarray}\label{targetABC}
m_\text{ABC}(\vtheta|{\bf y}_{\texttt{true}}, \epsilon)=  \int_{\mathbb{R}^{D_Y}} p_e(\vtheta,{\bf y}|{\bf y}_{\texttt{true}},\epsilon) d{\bf y} \propto \int_{\mathbb{R}^{D_Y}} h({\bf y}_{\texttt{true}}|{\bf y},\epsilon)\ell({\bf y}|\vtheta) g(\vtheta) d{\bf y}.  
\end{eqnarray}
The function $h({\bf y}_{\texttt{true}}|{\bf y},\epsilon)$ must be chosen such that $m_\text{ABC}(\vtheta|{\bf y}_{\texttt{true}}, \epsilon)$ converges to $p(\vtheta| {\bf y}_{\texttt{true}})$ as $\epsilon \rightarrow 0$.
Several computational algorithms designed for the ABC context are based on the {\it  noisy naive Monte Carlo} scheme in the extended space with target pdf $m_{ABC}(\vtheta|{\bf y}_{\texttt{true}},\epsilon)$ \footnote{\footnotesize Note that $\{\vtheta_t,{\bf y}_{t}^{(n)}\} \sim q(\vtheta,{\bf y})$  for all $n$. See the generalized chain rule in \cite{martino2018recycling}.} in Eq. \eqref{targetABC}, and proposal density $q(\vtheta,{\bf y})=\ell({\bf y}|\vtheta) g(\vtheta)$, { see Algorithm \ref{tab_noisy_IS_for_ABC}}.

\begin{algorithm}[!h]
	\centering
	\caption{Noisy importance sampling (N-IS) algorithm for ABC.}
	{\footnotesize
		\begin{tabular}{|p{0.95\columnwidth}|}
			\hline
			\begin{itemize}
				\item[1.] \textbf{Inputs:} {Prior} proposal distribution  $g(\vtheta)$ {(namely, $q(\vtheta)=g(\vtheta)$)}.
				
				\item[2.]For $t=1,\dots,T$:
				\begin{enumerate}
					\item[(a)]  Draw  $\vtheta_t\sim g(\vtheta)$,

					\item[(b)] Draw $M$ artificial data, ${\bf y}_{t}^{(1)},\ldots,{\bf y}_{t}^{(M)} \sim  \ell({\bf y}|\vtheta_t)$.

					\item[(c)] Assign to $\vtheta_t$, the noisy evaluation 
		\begin{equation}\label{Mtheta_Eq}
		\widetilde{p}_M(\vtheta_t)=\frac{1}{M}\sum_{i=1}^M h({\bf y}_{\texttt{true}}|{\bf y}_{t}^{(i)},\epsilon).
		\end{equation}

				\end{enumerate}	
			

				\item[3] \textbf{Outputs:}  Return $\{\vtheta_t,\widetilde{p}_M( \vtheta_t)\}_{t=1}^T$.
			\end{itemize}\\
			\hline
		\end{tabular}	
	}
	\label{tab_noisy_IS_for_ABC}
	\vspace{0.2cm}
\end{algorithm}

Thus, the pairs $\{\vtheta_t,\widetilde{p}_M(\vtheta_t)\}$ can be used for performing inference on $m_\text{ABC}(\vtheta|{\bf y}_\text{true},\epsilon)$.
Indeed, by standard Monte Carlo arguments, $\widetilde{p}_M(\vtheta) \approx  \int_{\mathbb{R}^{D_Y}} h({\bf y}_{\texttt{true}}|{\bf y},\epsilon)\ell({\bf y}|\vtheta) g(\vtheta) d{\bf y}$.
Increasing $M$, we reduce the variance of $\widetilde{p}_\epsilon(\vtheta)$, becoming closer and closer to $m_{ABC}(\vtheta|{\bf y}_{\texttt{true}}, \epsilon)$. Decreasing $\epsilon \rightarrow 0$, we reduce the bias between $m_{ABC}(\vtheta|{\bf y}_{\texttt{true}}, \epsilon)$ and $p(\vtheta|{\bf y}_{\texttt{true}})$, i.e.,{ the bias  $\mu(\vtheta,\epsilon) \rightarrow 0$  as $\epsilon \rightarrow 0$ (in this application context, the bias  $\mu(\vtheta,\epsilon)$ does not depend on $M$)}.
Instead of sampling $\vtheta_t$ from $g(\vtheta)$, we can use a generic proposal $q(\vtheta)$ (i.e., $q({\bf y},\vtheta)=\ell({\bf y}|\vtheta)q(\vtheta)$) and we obtain
\begin{equation}
	\widetilde{p}_M(\vtheta_t)=\left[\frac{1}{M}\sum_{i=1}^M h({\bf y}_{\texttt{true}}|{\bf y}_{t}^{(i)},\epsilon)\right] \frac{g(\vtheta_t)}{q(\vtheta_t)}, \qquad \theta_t \sim q(\vtheta).
\end{equation}
{
{\rem
For a fixed computational cost, there exists a trade-off between exploration and accuracy, i.e., between $T$ and $M$. For a related discussion, see \cite{doucet2015efficient,ko2021optimal,tran2013importance} and Section \ref{TradeoffSect}.
}}
\newline
\newline
Since simulating $M$ datasets for each $\vtheta$ can be costly, it has been proposed to use surrogates in order to accelerate the ABC algorithms. For instance, we can build a surrogate $\widehat{p}(\vtheta)$ considering the pairs $\{\vtheta_t,\widetilde{p}_M( \vtheta_t)\}$ or some related evaluations. 
In \cite{wilkinson2014accelerating}, a two-stage approach is used, where a GP surrogate of $\log m_\text{ABC}$ is built offline, and then a random-walk MH algorithm is applied on this surrogate 
An iterative refinement scheme using simulations $(\vtheta_t,{\bf y}_t^{(i)})$ is considered in \cite{meeds2014gps}.
Finally, the work by \cite{Gutmann15} combines Bayesian optimization with ABC in a two-stage scheme to build a surrogate of the discrepancy function $\Delta_\vtheta$ which measures the difference between ${\bf y}_\text{true}$ and ${\bf y}_\vtheta$, the data generated with parameter $\vtheta$. 
{
{\rem Note that, in the ABC framework, we have different possibilities when building the surrogate. 
Rather than building a surrogate of the (log) likelihood, we can model the discrepancy $\Delta_{\vtheta}$.
The advantages of modeling $\Delta_{\vtheta}$ are discussed in \cite{Gutmann15,jarvenpaa2019efficient}, for instance, $\Delta_{\vtheta}$ is independent of the bandwidth choice (and, more generally, of the kernel $h(\y_\text{obs}|\y,\vtheta,\epsilon)$). 
}
}

{\rem
In the ABC context, we can identify two surrogate functions: an internal surrogate $h({\bf y}_{\texttt{true}}|{\bf y},\vtheta,\epsilon)$ (that, generally, could also depends on $\vtheta$ as in the synthetic likelihood approach \cite{price2018bayesian}) and the external surrogate $\widehat{p}(\vtheta)$, for accelerating the algorithm.  
}


\subsection{Application to reinforcement learning}\label{RLsect}



Reinforcement learning (RL), which has many connections with control theory \cite{gullapalli1992reinforcement,sutton2018reinforcement}, is a popular and fast-growing area of machine learning.  An agent interacts with an environment by taking an action and, as a result of this action, it receives a state/observation and a reward. This occurs at each time step. 
One interaction/step is summarized as a state-action-reward triplet, $(s_t,a_t,r_t)$, where $t$ denotes the time index.
Therefore, an episode consists of $T$ steps over the environment (e.g., playing a game, if the environment represents a game, or otherwise interacting with the environment -- such as in robotics)
\begin{align}
\label{eq:tau}
\vtau &= \{s_0, (s_1,a_1,r_1),(s_2,a_2,r_2),\ldots,(s_T,a_T,r_T)\}=\{s_{0:T},a_{1:T},r_{1:T}\}.
\end{align}
The dynamics of the environment 
can be represented as follows, in the case of Markovian processes. For $t=1,2,\ldots,T$: 
\begin{gather}\label{StateModelSpace}
\left\{
\begin{split}
	&a_t  \sim \pi_{\vtheta}(a|s_{t-1}),\\
&s_{t} \sim p_\textsf{env}(s|s_{t-1},a_{t}), \\
&r_t \sim r_\textsf{env}(r|s_t,a_t,s_{t-1}),
\end{split}
\right.
\end{gather}
where the reward function $r_\textsf{env}$ and the transition function $p_\textsf{env}$ are determined by the application/environment. The policy $\pi_\vtheta(\cdot)$ determines which action the agent takes. Deterministic rules can be also employed for deciding $a_t$ and receiving a reward $r_t$. 
%
The payoff (i.e., accumulated reward, known as the return, or gain) for each episode is 
\begin{equation}
\label{eq:R}
R(\vtheta;\vtau) = \sum_{t=1}^{T} r_t.
\end{equation}
In certain settings, one can control the length $T$ of the episode $\vtau$. The goal is to find an optimal policy (i.e., optimal $\vtheta$) that maximizes the expected cumulative reward. There are a plethora of approaches to reinforcement learning, many falling under the category of so-called value-based methods (see \cite{sutton2018reinforcement} for an introduction and overview). 
Here, however, we focus specifically on the area of direct policy search, which is particularly apt for applications with continuous and small-but-complex action spaces such as robotics \cite{deisenroth2013survey}, and possibly non-Markovian settings (we refer to $p_{\mathsf{env}}$). More specifically, we focus on model-free policy search, i.e., learning the policy based on sampling trajectories; we do not attempt to recover $p_\textsf{env}$ or $r_\textsf{env}$. In this sense also, we are close to the large area of stochastic optimization \cite{PowellStochasticOptimization}. {Let us consider a prior $g(\vtheta)$}, we are interested in studying the following {pseudo-posterior} in the parameter space,
\begin{equation}
\label{eq:argmax}
 p(\vtheta) = {g(\vtheta)}\Exp_{\vtau} [R(\vtheta;\vtau)]=  \int_{\mathcal{T}} R(\vtheta;\bm{\tau})p(\bm{\tau}|\vtheta)d\bm{\tau},
\end{equation}
where $R(\cdot)$ from \Eq{eq:R}, and $\vtau \sim p(\bm{\tau}|\vtheta)$ is generated following the model in Eq. \eqref{StateModelSpace}, i.e.,
\begin{eqnarray}
&&p(\bm{\tau}|\vtheta) = p(s_{0:T},a_{0:T}, r_{1:T}|\vtheta), \nonumber \\
&&= p_0(s_0)\prod_{t=1}^Tr_\textsf{env}(r_t|s_t,a_t,s_{t-1}) p_\text{env}(s_t|s_{t-1},a_{t-1})\pi_\vtheta(a_{t-1}|s_{t-1}).
\end{eqnarray}
{ Note that $\Exp_{\vtau} [R(\vtheta;\vtau)]$ plays the role of an intractable pseudo-likelihood function. If the expected reward is proportional to a density w.r.t. $\vtheta$, we can consider a uniform prior $g(\vtheta) \propto 1$ and work directly on $p(\vtheta) = \Exp_{\vtau} [R(\vtheta;\vtau)]$.}
\newline
In a model-free direct search, we are not able to evaluate the distribution $p(\bm{\tau}|\vtheta)$, but we can draw from it by ``playing the game''.  Namely, we can estimate $p(\vtheta)$ by using sampled episodes. Given $M$ episodes $\bm{\tau}_{i} \sim p(\bm{\tau}|\vtheta)$ ($i=1,\dots,M$) generated according to $p(\bm{\tau}|\vtheta)$ with fixed $\vtheta$ (and fixing $T$), we can obtain the Monte Carlo estimation of the expected return
\begin{align}\label{EqRLimp}
\widetilde{p}_M( \vtheta) &= \frac{1}{N}\sum_{i=1}^M R(\vtheta; \bm{\tau}_{i}), \qquad \bm{\tau}_{i}\sim p(\bm{\tau}|\vtheta), \\
&= \frac{1}{M}\sum_{i=1}^M \sum_{t=1}^T r_t^{(i)}.
\end{align}
In this case, we have $m(\vtheta) = \Exp[\widetilde{p}_M( \vtheta)]=p(\vtheta)$ {(i.e., $\mu_M(\vtheta)=0$)}.
The variance 
\begin{eqnarray}
s^2(\vtheta)=\texttt{var}\left[\widetilde{p}_M( \vtheta) \right]=\frac{1}{N} \mbox{var}\left[R(\vtheta; \bm{\tau}_{i})\right].
\end{eqnarray}
The term $\mbox{var}[R(\vtheta,\vtau)]$ can have different forms depending on multiple aspects. 
The magnitude of the noise is reduced by averaging multiple episodes since the variance $s^2(\vtheta)$ decreases at rate $\frac{1}{M}$.
{\rem
As in the ABC setting, there is clearly a trade-off between precision in the evaluation of the target function and overall computational cost (which increases as $N$ grows).
This trade-off has been studied in the context of MCMC and IS \cite{doucet2015efficient,ko2021optimal}.
}
\newline
\newline
Note that the distribution $\widetilde{p}_M( \vtheta)$ also depends of the length $T$ of the episode. More specifically, the variance of the random variable $\widetilde{p}_M( \vtheta)$ decreases with $T$. 
If the process is ergodic, averaging over very long periods is equivalent to repeating the process multiple times. The noise can therefore be reduced by both prolonged simulation or repeated sampling at the expense of a higher computational cost per function evaluation.

\section{Numerical experiments}\label{numexperimentSect}  

In this section, we compare different algorithms discussed in Section 4.
It is important to remark that all the techniques are always compared with the same number of evaluations (denoted as $E$) of the noisy target pdf. Moreover, a  k-nearest neighbor (kNN) regression is applied in order to construct the surrogate function. Recall that the baseline PM-MH algorithm is not using a surrogate model (see Algorithm \ref{tab_pseudoMH}). 
\newline
In the first experiment, the target is a two-dimensional banana-shaped density which is non-linear benchmark in the literature \cite{cornuet2012adaptive}, perturbed with two different noises: one is an unbiased noise, and with the other noisy the target distribution becomes a heavy-tailed banana pdf.
The second experiment considers a multimodal target density.  
Finally, we apply the algorithms in a benchmark RL problem consisting on balancing two poles attached to a cart.


\subsection{Non-linear banana density}\label{sec_banana}

We consider a banana-shaped target pdf,
\begin{align}\label{eq:BananaTarget}
	p(\vtheta) \propto \exp \left( -\frac{(\eta_1-B\theta_1-\theta_2^2)^2}{2\eta_0^2}  - \frac{\theta_1^2}{2\eta_{1}^2}
	-\frac{\theta_2^2}{2\eta_{2}^2}
	\right),
\end{align}
with $B=4$, $\eta_0=4$ and $\eta_i= 3.5$ for $i=1,\ldots,2$, where $\Theta = [-10,10]\times[-10,10]$, i.e., bounded domain. The goal is to compare the performance of the different algorithms against a vanilla PM-MH algorithm for two different noises. Specifically, we compare 
\begin{itemize}
	\item {\bf (1)} DA-PM-MH with $T_\text{surr}=1$,

	\item {\bf (2)} DA-PM-MH with $T_\text{surr}=5$,
	
	\item {\bf (3)} MH-S with $\rho_\text{update}=1$,

	\item {\bf (4)} MH-S with $\rho_\text{update}=\alpha_\text{MH}$. 
\end{itemize}
The baseline corresponds to a PM-MH algorithm with $5000$ iterations. 
We consider the same proposal $\varphi(\vtheta|\vtheta')=\mathcal{N}(\vtheta|\vtheta',3^2\bm{I}_2)$
for all the methods (including the baseline).
The surrogate is built with  k-nearest neighbor (kNN) regression using $K\in\{1,10,100\}$ neighbors. 
For all methods, the surrogate is initialized as a uniform distribution and updated from there on using the incoming realizations $\widetilde{p}_M( \vtheta)$. 
Note that MH-S with $\rho=1$ is equivalent to PM-MH when $K=1$. We include it in that case for the sake of completeness.
We set $E=5000$ as the budget of noisy target evaluations. 
\newline
In addition, we have applied IS schemes for the two noises. Specifically, we compare standard (noisy) IS against N-DIS, using again the nearest neighbor surrogate. 
For the standard noisy IS, we use a uniform proposal in $\mathcal{X}$.
For N-DIS, we test $T=5,N=1000$ and $T=10,N=500$, so that the total number of evaluations is $E=NT=5000$.
\newline
\textbf{Unbiased banana.}
First, we consider the noise $\widetilde{p}_M( \vtheta) = \epsilon p(\vtheta)$ with $\epsilon \sim \mbox{Exp}(1)$. 
In this case, the expected target is $p(\vtheta)$. 
We consider the estimation of the mean and the diagonal of the covariance matrix, whose ground truths are $\bm{\mu} = [ -0.48,0]$ and  $\text{diag}(\bm{\Sigma})=[1.38, 8.90]$.
We show the results in Figures \ref{fig:bananaRuidoExp} and \ref{fig:bananaRuidoExpIS}.
\newline
\textbf{Heavy-tailed banana.}
Then, we consider the noise $\widetilde{p}_M( \vtheta) = \max(0,p(\vtheta)+\epsilon)$
with $\epsilon\sim \mathcal{N}(0,0.01^2)$.
For this choice, we have $m(\vtheta) \neq p(\vtheta)$, so we have to evaluate the performance in the estimation of the new moments, i.e., this noise changes the density that the methods target, whose ground truths are $\widetilde{\bm{\mu}} = [ -0.38, 0]$ and 
$\text{diag}(\widetilde{\bm{\Sigma}}) = [6.74, 12.84]$. 
The resulting density $m(\vtheta)$ has constant tails since this noise introduce bias in the low probability regions (as in Figure \ref{fig_target_ruidosa_1D} in App. \ref{illustrative_ex_1D}). We show the results in Figures \ref{fig:bananaRuidoMax} and \ref{fig:bananaRuidoMAXIS}.

\subsubsection{Dependence on the surrogate}

\begin{figure}[h!]
	\centering
	\begin{subfigure}[b]{0.48\textwidth}
		\includegraphics[width=1\textwidth]{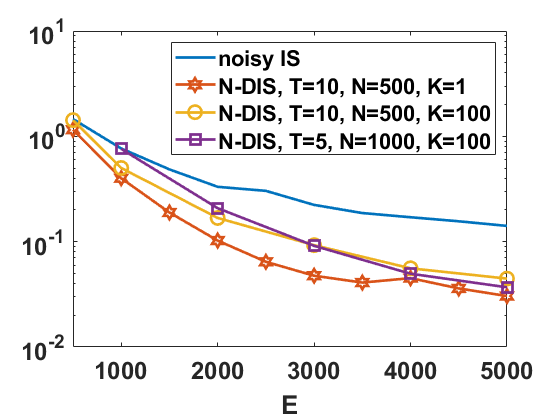}
		\caption{Estimation of the mean.}
	\end{subfigure}
	\begin{subfigure}[b]{0.48\textwidth}
		\includegraphics[width=1\textwidth]{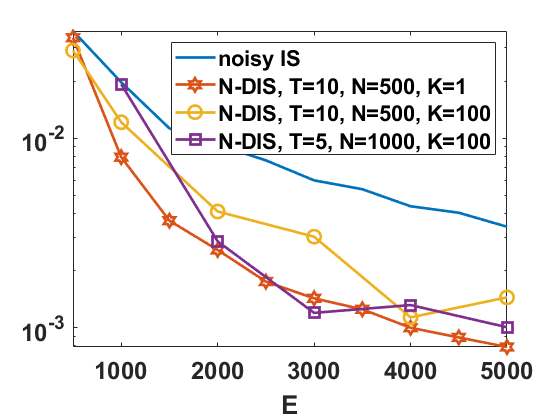}
		\caption{Estimation of the variance.}
	\end{subfigure}
	
	\caption{\label{fig:bananaRuidoExpIS}
		Relative median squared error in estimation of the mean (left) and variance (right) of the banana pdf with multiplicative exponential noise, by importance sampling schemes.	
	}
\end{figure}

\begin{figure}[h!]
	\centering
	\begin{subfigure}[b]{0.48\textwidth}
		\includegraphics[width=1\textwidth]{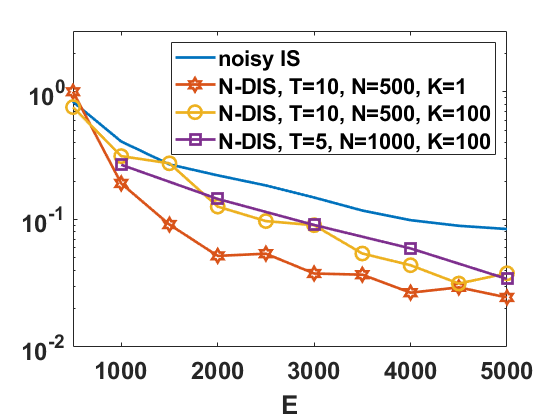}
		\caption{Estimation of the mean.}
	\end{subfigure}
	\begin{subfigure}[b]{0.48\textwidth}
		\includegraphics[width=1\textwidth]{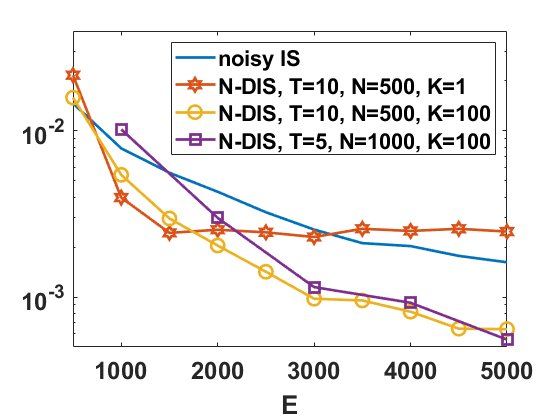}
		\caption{Estimation of the variance.}
	\end{subfigure}
	
	\caption{\label{fig:bananaRuidoMAXIS}
		Relative median squared error in estimation of the mean (left) and variance (right) of the banana pdf with rectified additive Gaussian noise, by importance sampling schemes.	
	}
\end{figure}

The use of surrogate improves the performance, but can be detrimental as well.
This duality accounts for the differences in performance between estimating $\bm{\mu}$ (upper rows of Figures \ref{fig:bananaRuidoExp} and \ref{fig:bananaRuidoMax}) and estimating $\text{diag}(\bm{\Sigma})$ (lower rows of Figures \ref{fig:bananaRuidoExp} and \ref{fig:bananaRuidoMax}).
\newline
{\bf Benefits of using surrogates.}
For both noises, the considered algorithms perform better than the baseline in the estimation of $\bm{\mu}$ for all $K$,
something that it is related to properly visiting the regions of high probability. In this sense, it shows that using surrogates within MCMC algorithms help in discovering high-probability regions.
In IS, the use of surrogates also improves the performance in the estimation of the mean, as it can be seen in {Figure \ref{fig:bananaRuidoExpIS}(a)} and Figure \ref{fig:bananaRuidoMAXIS}(a).  
\newline
{\bf Pathological constructions.}
Both choices of noise produce noisy realizations $\widetilde{p}_M( \vtheta)$ that are skewed towards 0, specially in the low-probability regions.
A surrogate built with such evaluations may difficult the exploration of the tails of the distribution.
This can be seen at the error in estimating the variance in Figure \ref{fig:bananaRuidoExp}(d) and Figure \ref{fig:bananaRuidoMax}(d), where the considered methods perform worse than the baseline.
Although the DA-PM-MH algorithms (with $T_\text{surr}=1$ and $T_\text{surr}=5$) are ``exact'', they fail at estimating the variance since the surrogate does not fulfill the minimum requirements. In fact, a `bad' surrogate is preventing the chain to explore the regions properly.
Increasing $K$ makes the surrogate smoother and hence should improve the variance estimation. This is confirmed in Figure \ref{fig:bananaRuidoExp}(e)-(f) and Figure \ref{fig:bananaRuidoMax}(e)-(f), where the DA-PM-MH  algorithms perform better than the baseline.
The MH-S algorithms present a trade-off between performance and exactness/bias as we increase K, that we comment below.
{Similarly, we see in Figure \ref{fig:bananaRuidoMAXIS}(b) that N-DIS with $K=1$ also have difficulties converging in estimating the variance of the heavy-tailed banana, due to a bad surrogate that prevents visiting the tails.}

\subsubsection{Bias in iterative refinement algorithms}
\begin{figure}[h!]
	\centering
	\begin{subfigure}[b]{0.31\textwidth}
		\includegraphics[width=1\textwidth]{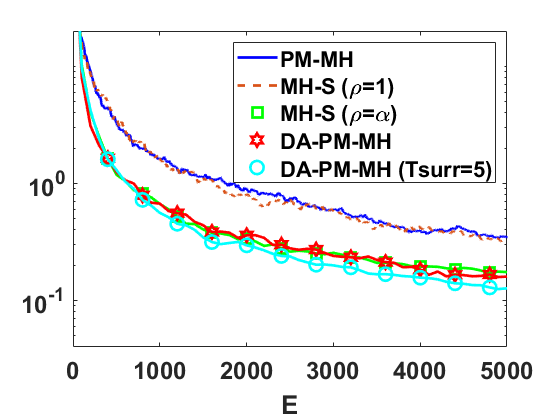}
		\caption{$K=1$}
	\end{subfigure}
	\begin{subfigure}[b]{0.31\textwidth}
		\includegraphics[width=1\textwidth]{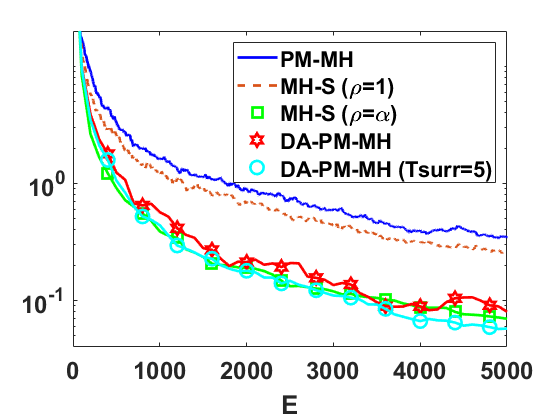}
		\caption{$K=10$}
	\end{subfigure}
	\begin{subfigure}[b]{0.31\textwidth}
		\includegraphics[width=1\textwidth]{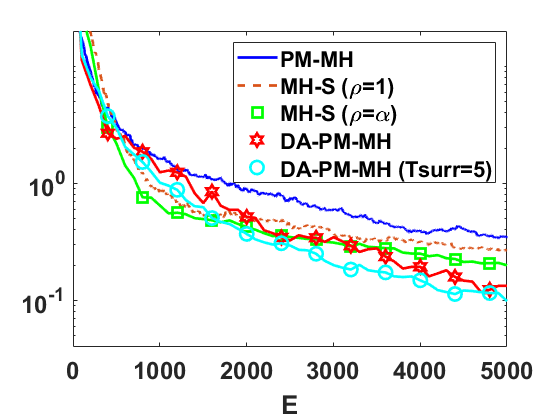}
		\caption{$K=100$}
	\end{subfigure}

	\begin{subfigure}[b]{0.31\textwidth}
		\includegraphics[width=1\textwidth]{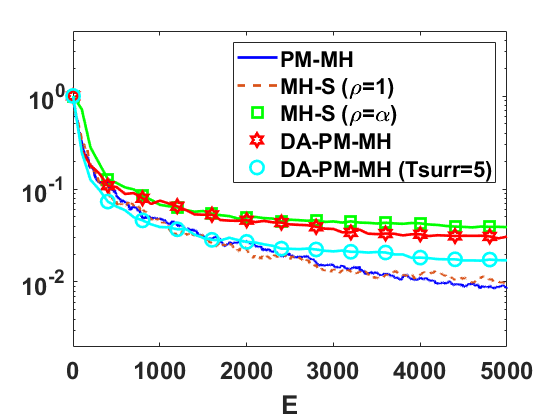}
		\caption{$K=1$}
	\end{subfigure}
	\begin{subfigure}[b]{0.31\textwidth}
		\includegraphics[width=1\textwidth]{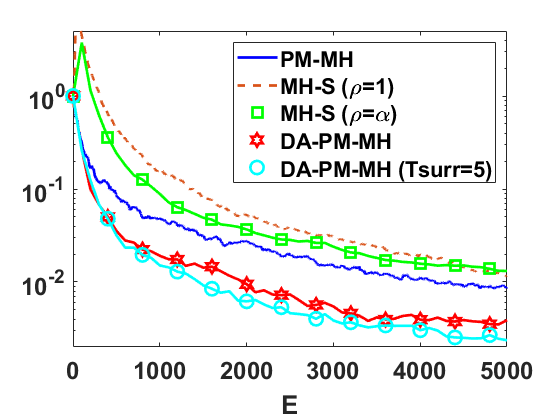}
		\caption{$K=10$}
	\end{subfigure}
	\begin{subfigure}[b]{0.31\textwidth}
		\includegraphics[width=1\textwidth]{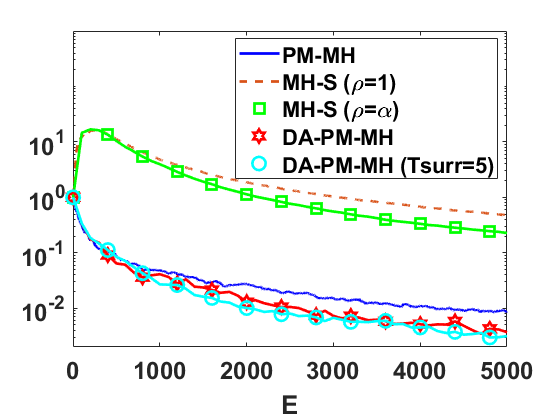}
		\caption{$K=100$}
	\end{subfigure}
	\caption{\label{fig:bananaRuidoExp}
		Relative median squared error in estimation of the mean (upper row) and variance (lower row) of the banana pdf with multiplicative exponential noise.	
	}
\end{figure}

\begin{figure}[h!]
	\centering
	\begin{subfigure}[b]{0.31\textwidth}
		\includegraphics[width=1\textwidth]{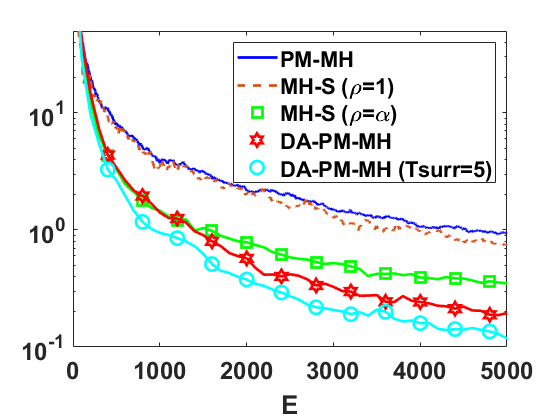}
		\caption{$K=1$}
	\end{subfigure}
	\begin{subfigure}[b]{0.31\textwidth}
		\includegraphics[width=1\textwidth]{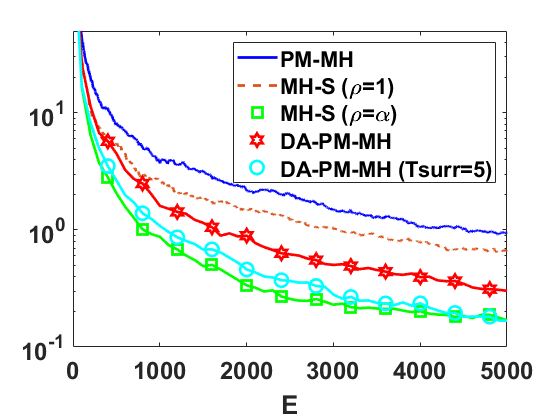}
		\caption{$K=10$}
	\end{subfigure}
	\begin{subfigure}[b]{0.31\textwidth}
		\includegraphics[width=1\textwidth]{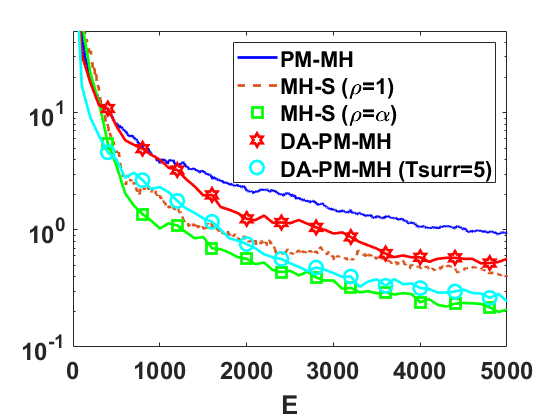}
		\caption{$K=100$}
	\end{subfigure}

	\begin{subfigure}[b]{0.31\textwidth}
		\includegraphics[width=1\textwidth]{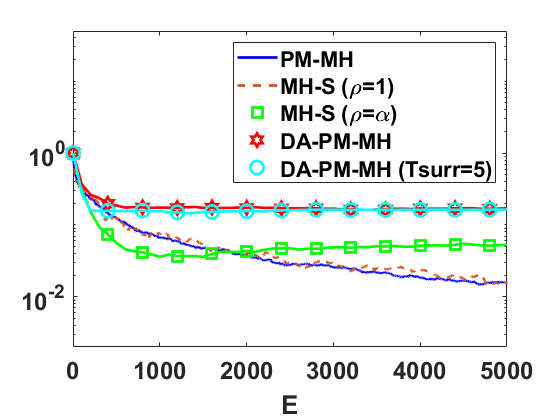}
		\caption{$K=1$}
	\end{subfigure}
	\begin{subfigure}[b]{0.31\textwidth}
		\includegraphics[width=1\textwidth]{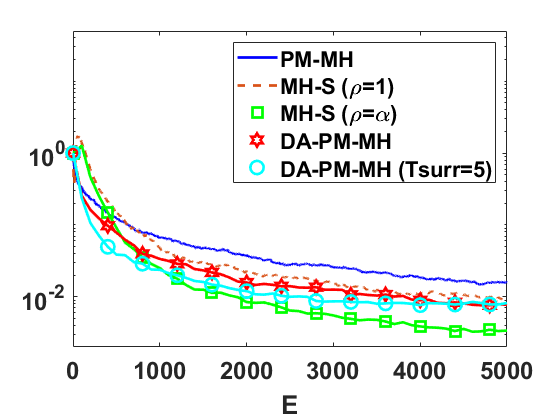}
		\caption{$K=10$}
	\end{subfigure}
	\begin{subfigure}[b]{0.31\textwidth}
		\includegraphics[width=1\textwidth]{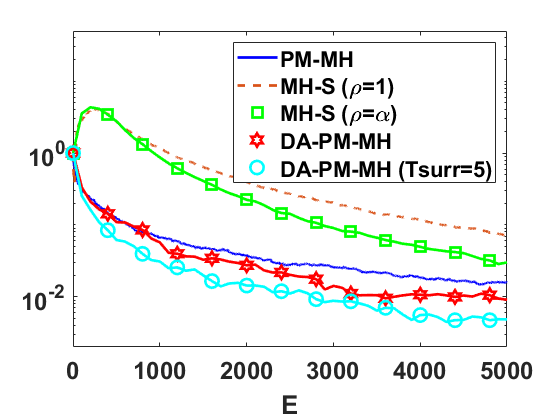}
		\caption{$K=100$}
	\end{subfigure}
	\caption{\label{fig:bananaRuidoMax}
		Relative median squared error in estimation of the mean (upper row) and variance (lower row) of the banana pdf perturbed as $\fapprox(\vtheta) = \max(0,p(\vtheta)+\epsilon), \ \epsilon \sim \mathcal{N}(0,0.01)$.	
	}
\end{figure}

%
Since these algorithms target the surrogate, the choice of $K$ affects the performance. 
In Figures \ref{fig:bananaRuidoExp}(a)-(c), we see the algorithms MH-S with $\rho=1$ and $\rho=\alpha$ beat the baseline in the estimation of $\bm{\mu}$. 
However, in Figures \ref{fig:bananaRuidoExp}(d)-(f) the situation is the opposite, performing worse than the baseline in the estimation of $\mbox{diag}(\bm{\Sigma})$ for the $K$ considered. 
As we commented above, the exponential distribution with $\lambda=1$ concentrates around 0, hence this noise tends to give noisy realizations that underestimate the true density.  In low-probability regions and when $K=1$, this phenomenon amplifies since realizations with very low value difficult that their neighborhood gets properly explored. This is why MH-S is able to estimate $\bm{\mu}$ with $K=1$ (i.e. the high-probability region is properly visited), but fails at estimating $\mbox{diag}(\bm{\Sigma})$.
\newline
We increase $K$ in order to reduce this problem. However, attending to Figures \ref{fig:bananaRuidoExp}(e)-(f) for $K=10$ and $K=100$, both MH-S still perform poorly in the estimation of $\mbox{diag}(\bm{\Sigma})$. Now, this is because the surrogate has huge bias (since, for fixed number of nodes, as we consider more neighbors, the surrogate becomes a flattened version of $p(\vtheta)$).
In other words, regarding the choice of $K$ for the MH-S, the increase in performance is traded off with exactness. 
Note that this bias is detected when estimating the variance, since this biased surrogate has $\bm{\mu}$ almost unaltered.
\newline
Regarding the second type of noise in Figure \ref{fig:bananaRuidoMax}, the conclusions are similar.
In Figure \ref{fig:bananaRuidoMax}(d), we see that estimation of the variance is even worse with this second noise, since the target has now constant tails which are not captured by the surrogate with $K=1$. 
However, MH-S algorithms perform better (w.r.t.\ the previous noise) in the estimation of the variance for $K=10$. This is probably due to the surrogate having a low bias w.r.t.\ the true target $m(\vtheta)$, which is broader than in the previous noise.


%

\subsection{Bimodal target density}

\begin{figure}[h!]
	\centering
	\begin{subfigure}[b]{0.31\textwidth}
		\includegraphics[width=1\textwidth]{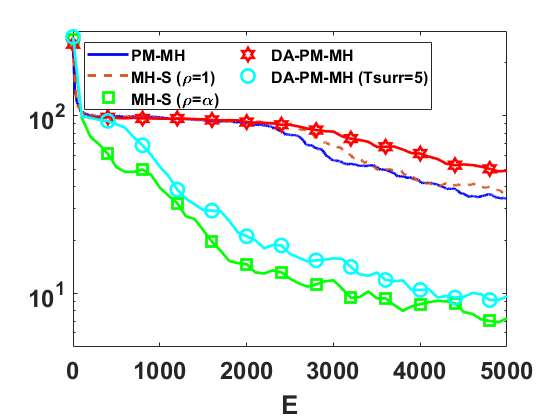}
		\caption{$K=1$}
	\end{subfigure}
	\begin{subfigure}[b]{0.31\textwidth}
		\includegraphics[width=1\textwidth]{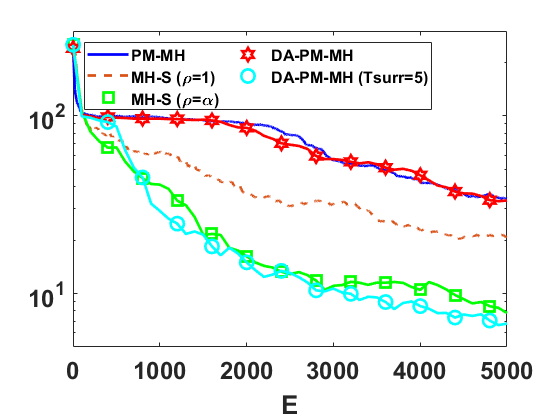}
		\caption{$K=10$}
	\end{subfigure}
	\begin{subfigure}[b]{0.31\textwidth}
		\includegraphics[width=1\textwidth]{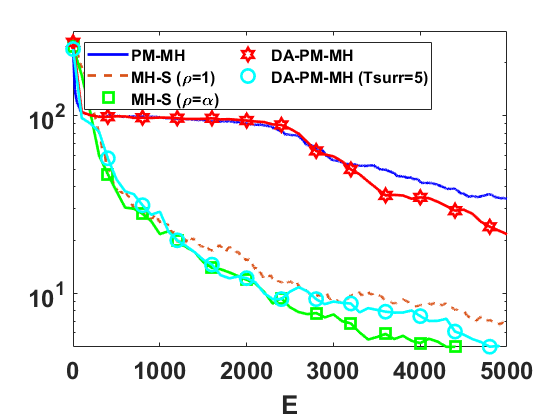}
		\caption{$K=100$}
	\end{subfigure}

	\begin{subfigure}[b]{0.31\textwidth}
		\includegraphics[width=1\textwidth]{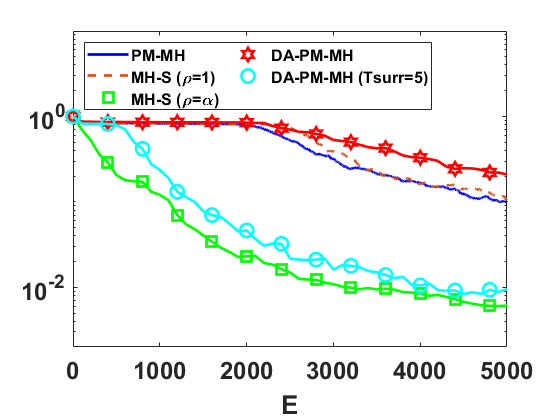}
		\caption{$K=1$}
	\end{subfigure}
	\begin{subfigure}[b]{0.31\textwidth}
		\includegraphics[width=1\textwidth]{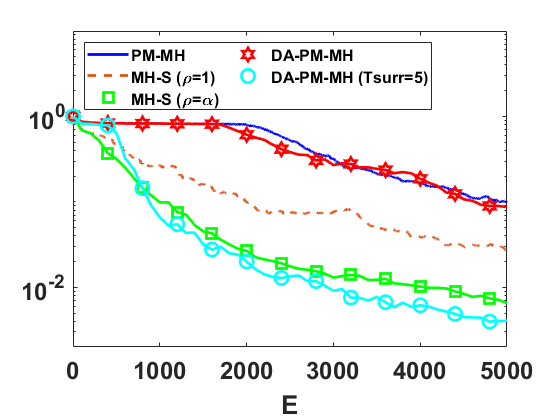}
		\caption{$K=10$}
	\end{subfigure}
	\begin{subfigure}[b]{0.31\textwidth}
		\includegraphics[width=1\textwidth]{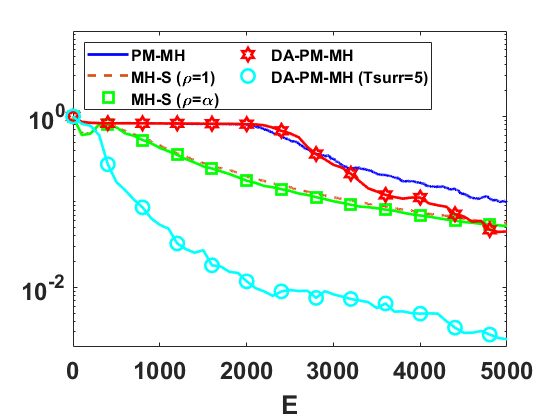}
		\caption{$K=100$}
	\end{subfigure}
	\caption{\label{fig:bimodalRuidoExp}
		Relative median squared error in estimation of the mean (upper row) and variance (lower row) of the bimodal pdf with multiplicative exponential noise.	
	}
\end{figure}

\begin{figure}[h!]
	\centering
	\begin{subfigure}[b]{0.48\textwidth}
		\includegraphics[width=1\textwidth]{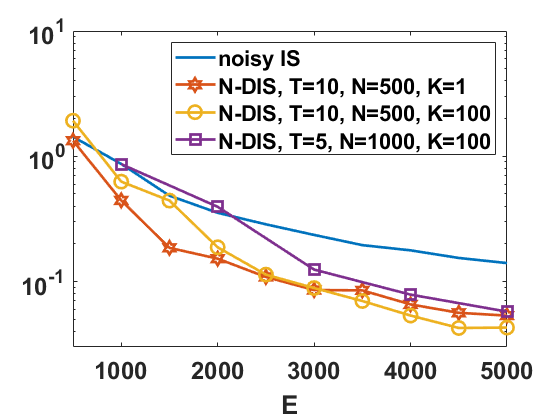}
		\caption{Estimation of the mean.}
	\end{subfigure}
	\begin{subfigure}[b]{0.48\textwidth}
		\includegraphics[width=1\textwidth]{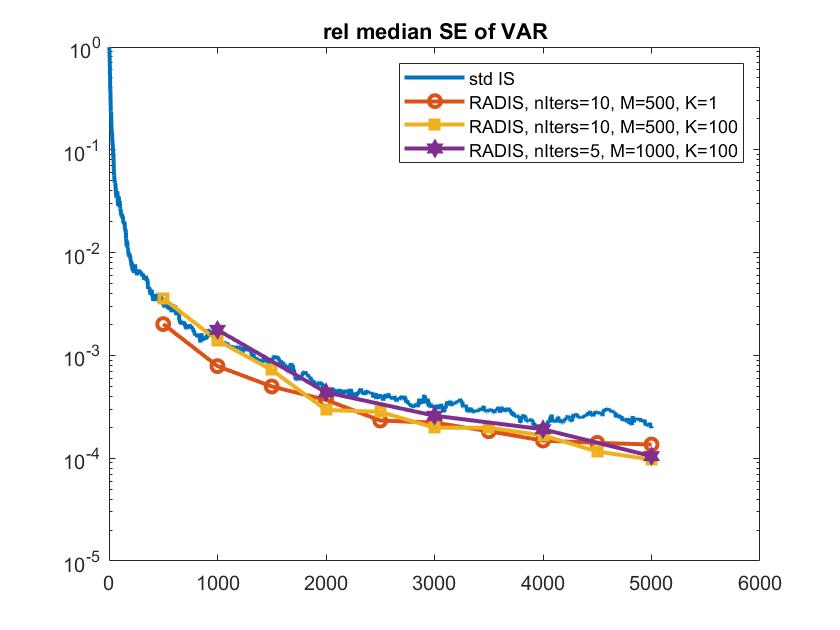}
		\caption{Estimation of the variance.}
	\end{subfigure}
	
	\caption{\label{fig:bimodalRuidoExpIS}
		Relative median squared error in estimation of the mean (left) and variance (right) of the bimodal pdf with multiplicative exponential noise, by importance sampling schemes.	
	}
\end{figure}

Now, we consider the density
$$
p(\vtheta) = \frac{1}{2}\mathcal{N}(\vtheta|[10,0]^\top,3^2\bm{I}_2)
+\frac{1}{2}\mathcal{N}(\vtheta|[-10,0]^\top,3^2\bm{I}_2),
$$
where $\Theta = [-20,20]\times[-20,20]$, i.e., bounded domain. 
We consider the noise $\widetilde{p}_M( \vtheta) = \epsilon p(\vtheta)$ with $\epsilon \sim Exp(1)$. 
As in the previous experiment, we compare the algorithms in the estimation of the mean $\bm{\mu} = [0,0]^\top$ and the diagonal of the covariance matrix $\mbox{diag}(\bm{\Sigma}) = [108.87, 9]^\top$. 
For the MCMC algorithms, this time we consider a proposal, $\varphi(\vtheta|\vtheta')=\mathcal{N}(\vtheta|\vtheta',2^2\bm{I}_2)$, intentionally chosen so that the mixing can be slow for some initializations.
We set $E=5000$ as the budget of noisy evaluations.
Results are shown in Figure \ref{fig:bimodalRuidoExp}. The results of the IS schemes on the same noisy target are shown in Figure \ref{fig:bimodalRuidoExpIS}. 
\newline
{\bf Improved exploration by surrogates.}
In this example, the chosen proposal $\varphi(\vtheta'|\vtheta)$ is not able to explore efficiently the space since the two modes are rather distant. 
For this reason, the results of PM-MH are much worse than the algorithms that perform several steps w.r.t.\ the surrogate, namely, DA-PM-MH with $T_\text{surr}=5$ and MH-S with $\rho=\alpha$, as can be seen in Figure \ref{fig:bimodalRuidoExp}.
This shows that performing several steps w.r.t.\ surrogate is beneficial for the exploration and for discovering different modes, specially when the proposal does not propose big jumps.
Regarding the results of IS, we see in Figure \ref{fig:bimodalRuidoExpIS} that the use of surrogates improve the performance, but not as much as in the MCMC test, since IS with uniform density already performs very well as compared to PM-MH.
\newline
{\bf Pathological constructions.}
In this example, we encounter the negative effect of a bad surrogate construction. In Figures \ref{fig:bimodalRuidoExp}(a)-(b)-(d)-(e),
we see that DA-PM-MH with $T_\text{surr}=1$ performs equal or worse than the baseline technique, i.e., PM-MH. This is probably due to the joint effect of small jumps proposed by $\varphi(\vtheta|\vtheta')$ and performing only one step w.r.t.\ the surrogate, which in turn makes a myopic construction of the surrogate possibly missing one of the modes.
This pathological behavior is worst when $K=1$, but improves as we increase $K$, matching the performance of PM-MH for $K=100$. 

%
%

\subsection{Double cart pole}\label{Double_cart_poleSect}

We consider a variant of the popular cart-pole system, which is a standard benchmark in RL \cite{heidrich2009neuroevolution}.
In the basic cart-pole environment, the goal is to balance a pole that is hinged on  a cart. The cart is able to move freely along the x-axis. 
The observations are the position $x$ and velocity $\dot{x}$ of the cart, and the angle $\alpha$ and angular velocity $\dot{\alpha}$ of the pole.
The action is continuous and corresponds to the force applied to the cart.
The agent receives one point for each iteration that $x$ and $\alpha$ are within some bounds.

\begin{figure}[h!]
	\centering
	\includegraphics[scale=0.45]{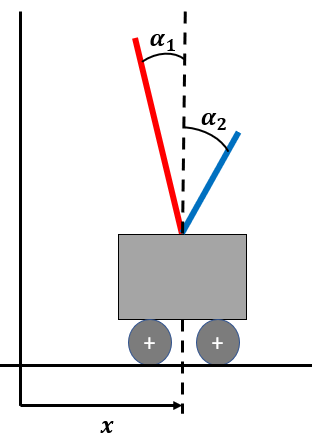}
	\caption{\label{fig_double_cartpole} Double pole balancing problem.}
\end{figure}

We consider here the more challenging variant where another shorter pole is hinged on the cart (see Figure \ref{fig_double_cartpole}). 
Hence, the state vector is ${\bf s} = [x,\dot{x},\alpha_1,\dot{\alpha}_1,\alpha_2,\dot{\alpha}_2]^\top$.
The transition $p_\text{env}$ is deterministic, determined by the evolution of the dynamical system, where each iteration corresponds to $0.02$s \cite{wieland1991evolving}.
We consider {a linear} policy $a = \pi_\vtheta({\bf s})=\vtheta^\top {\bf s}$. 
Hence, the parameter dimension is $\vtheta\in\mathbb{R}^6$.
The return $R(\vtheta,\vtau)$ is the number of iterations before any of $x$, $\alpha_1$ or $\alpha_2$ go out of bounds, where $T_\text{max}=1000$. Hence, the maximum return is 1000.
Regarding the parameters such as the masses, lengths, friction coefficients, etc., we take the same values as in \cite{heidrich2009neuroevolution}. 
At the beginning of each episode, the initial state is obtained by sampling each component uniformly within the following intervals: $x \in [-1.944, 1.944]$, $\dot{x} \in [-1.215, 1.215]$, $\alpha_1\in [-0.0472, 0.0472]$, $\dot{\alpha}_1\in[-0.135088, 0.135088]$, $\alpha_2 \in [-0.10472, 0.10472]$ and $\dot{\alpha}_2\in [-0.135088, 0.135088]$.
Note that, in this example, the noisiness comes only from the initial distribution.

\begin{figure}[h!]
	\centering
	\begin{subfigure}[b]{0.31\textwidth}
		\includegraphics[width=1\textwidth]{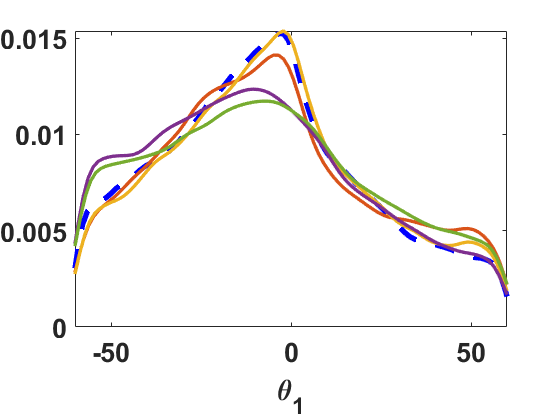}
	\end{subfigure}
	\begin{subfigure}[b]{0.31\textwidth}
		\includegraphics[width=1\textwidth]{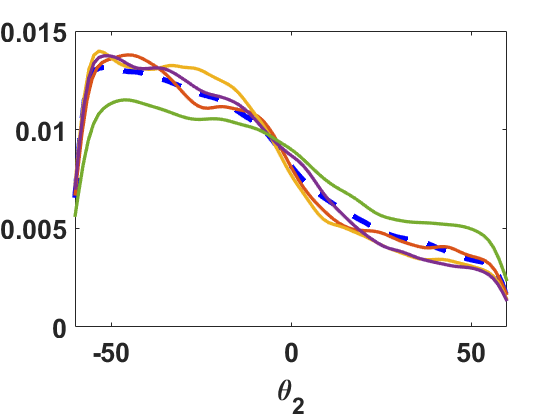}
	\end{subfigure}
	\begin{subfigure}[b]{0.31\textwidth}
		\includegraphics[width=1\textwidth]{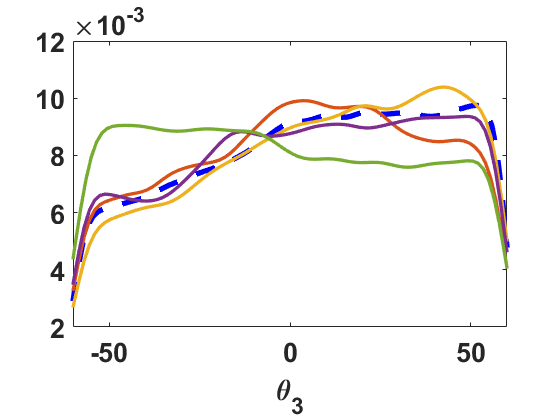}
	\end{subfigure}
	\begin{subfigure}[b]{0.31\textwidth}
	\includegraphics[width=1\textwidth]{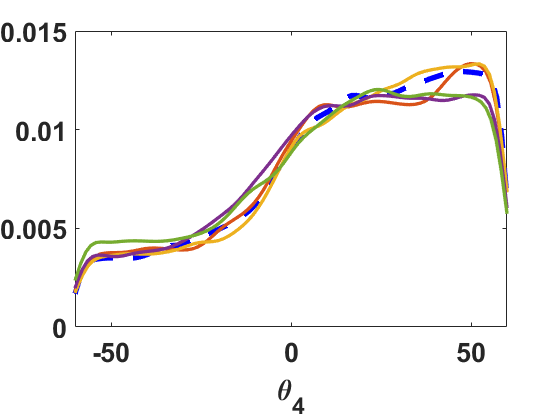}
\end{subfigure}
\begin{subfigure}[b]{0.31\textwidth}
	\includegraphics[width=1\textwidth]{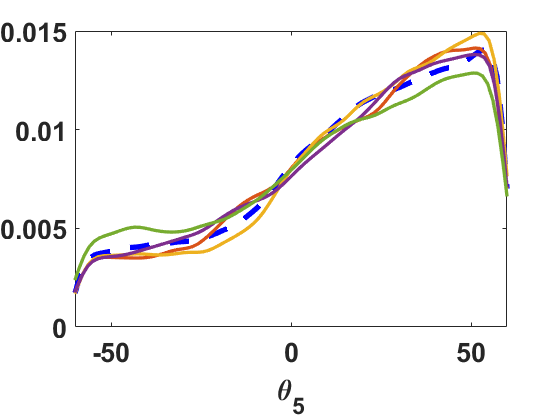}
\end{subfigure}
\begin{subfigure}[b]{0.31\textwidth}
	\includegraphics[width=1\textwidth]{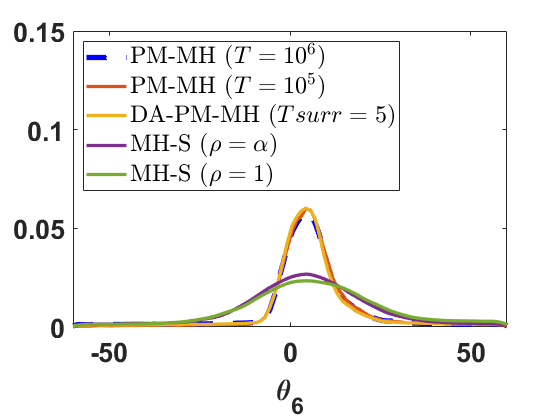}
\end{subfigure}
	\caption{\label{fig_double_cartpole_marginals} Marginal densities of the double cart pole system, obtained by the different algorithms.	
	}
\end{figure}

We consider a realization $\widetilde{p}_M( \vtheta)$ of $p(\vtheta)$ by simulating one single episode.
We first run $10^6$ iterations of PM-MH on $\widetilde{p}_M( \vtheta)$ in order to have a rough estimation (groundtruth) of the marginal histograms w.r.t.\ we can compare the algorithms. 
We consider a bounded domain $\vtheta\in [-60,60]^6$.
We compare two MH-S algorithms and one DA-PM-MH algorithm using again a nearest neighbor surrogate, with $K=100$. The budget is $E=10^5$ evaluations.
A PM-MH algorithm with the same number of evaluations is also considered.
In Figure \ref{fig_double_cartpole_marginals}, we show the estimated marginal densities.
In Table \ref{tab_double_cartpole_groundtruths}, we show the MMSE estimations of $\vtheta$ provided by the different algorithms. We can observe that the compared techniques are able to approximate the groundtruth marginal histograms. However, the DA-PM-MH scheme seems to provide slightly better approximations.

\begin{table}
	\begin{center}
		\scriptsize
		\begin{tabular}{| l | c | c | c | c | c | c | c |} 
			\hline
			& $\theta_1$ & $\theta_2$ & $\theta_3$ & $\theta_4$ & $\theta_5$ & $\theta_6$ & Exp. return\\ [0.5ex] 
			\hline\hline
			PM-MH ($T=10^6$) & -7.1281 & -15.0300  &  5.1756  & 15.0946 &  15.4696  &  4.9734 & 1000 \\
			PM-MH ($T=10^5$) & -5.6738 & -15.7544 &  3.0080 &  14.9182 &  16.3909  &  6.0570 & 1000 \\
			MH-S ($\rho=1$) & -6.6351 & -10.2346 &  -1.9859 &  12.5025 &  12.8274  &  6.0455 & 1000 \\
			MH-S ($\rho=\alpha$) & -8.9285 & -17.0432  &  4.0197 & 13.3249 & 15.7900  &  3.9512 & 1000\\ 
			DA-PM-MH ($T_\text{surr}=5$) & -5.7748 & -17.5469  &  6.6250 &  15.9932  & 17.5892  &  5.2058 & 1000 \\
			\hline
		\end{tabular}
		\caption{\label{tab_double_cartpole_groundtruths} MMSE estimates for the double cart pole system computed by the different algorithms.}
	\end{center}
	
\end{table}

{

\subsection{Application to the ABC framework}


We illustrate some of the concepts presented in this work within the specific setting of approximate Bayesian computation (ABC), namely the noisy realizations associated with pointwise evaluations of the likelihood, how this noisy can be controlled by the user at the expense of an increased computational cost, and the improved performance of Monte Carlo algorithms by the use of clever surrogate constructions. 
Let us consider $L=10$ observations $\y_\text{obs} = [y_1,\dots,y_L]$ drawn from a Gaussian distribution with mean $3$ and unit variance. We desire to make inference on the mean $\vtheta = \theta\in\R$ of the Gaussian likelihood $\mathcal{N}(\theta,1)$. Assuming also a Gaussian prior $\theta \sim g(\theta) = \mathcal{N}(\theta|\mu_0,\sigma_0)$ with $\mu_0=0$ and $\sigma_0=3$, the true posterior distribution is $p(\theta|\y) = \mathcal{N}(\theta|\mu_p,\sigma_p^2)$ with
$\mu_p =  \sigma_p^{2}(  \frac{\mu_0}{\sigma_0^2} + \sum_{i}^Ly_i )$ and $\sigma_p^2 = (\frac{1}{\sigma_0^2} + L )^{-1}$.
We assume that we do not have access to the likelihood function and consider the accept-reject ABC scheme, that uses  $h(\y_\text{obs}|\y,h,\theta) = \mathbbm{1}\left( \Delta_{\theta} < \epsilon \right)$, where $\Delta_{\theta} = \lvert \bar{\y}_\text{obs} - \bar{\y}_\theta)\rvert^2$, where $\bar{\y}_\text{obs} $, $\bar{\y}$ denote the sample mean and $\epsilon$ is the threshold.
 In this toy example, we can write down the analytical ABC posterior in Eq. \eqref{targetABC},
\begin{align*}
&m_\text{ABC}(\theta|\y_\text{obs}) \propto g(\theta)\ell_\text{ABC}(\y_\text{obs}|\theta), \\
\ell_\text{ABC}(\y_\text{obs}|\theta) &\propto \mathbb{P}(\Delta_{\theta}<\epsilon) = \Phi\left(  \sqrt{L}(\y_\text{obs} - \theta) + \sqrt{L\epsilon} \right) - \Phi\left(  \sqrt{L}(\y_\text{obs} - \theta) - \sqrt{L\epsilon} \right),
\end{align*}
where $\Phi(\cdot)$ denotes the cdf of the standard univariate Gaussian \cite{Gutmann15}. For fixed $\theta$, unbiased estimates of $\ell_\text{ABC}(\y_\text{obs}|\theta)$ can be obtained by sampling a set of synthetic datasets $\y^{(m)}_{\theta} \sim \mathcal{N}(\theta,1)$ for $m=1,\dots,M$, and computing 
$$
\ell_\text{ABC}(\y_\text{obs}|\theta) \approx \widetilde{\ell}_\text{ABC}(\y_\text{obs}|\theta) = \frac{1}{M}\sum_{m=1}^M \mathbbm{1}\left( \Delta^{(m)}_{\theta} < \epsilon \right),
$$
where $\Delta^{(m)}_{\theta} = \lvert {\bar \y}_\text{obs} - {\bar \y}^{(m)}_\theta\rvert^2$.
 Hence, noisy Monte Carlo algorithms can be applied to approximate $m_\text{ABC}(\theta|\y_\text{obs})$, namely, the pseudo-marginal MH algorithm and noisy IS algorithms (see Sect. \ref{sec_abc}). However, if $h$ is small, obtaining good estimates of the likelihood for some $\theta$ is very costly since most of the simulated datasets will have $\mathbbm{1}( \Delta^{(m)}_{\theta} < \epsilon ) = 0$. Then, we aim to apply regression techniques to approximate $\ell_\text{ABC}(\y_\text{obs}|\theta)$ from a set of noisy evaluations.  
\newline
\newline
The noise in the evaluation of ABC likelihood is controlled by $M$, the number of synthetic datasets for each $\theta$. 
In the standard ABC setting, a single dataset ($M=1$) is employed to approximate the evaluation of $\ell_\text{ABC}(\y_\text{obs}|\theta)$.
In Figure \ref{fig_abc_noisy_lik}(a)-(c), we show that better approximations can be obtained with larger values $M$ at the expense of increasing the computational cost. 
It is important to notice that the noiseless $m_\text{ABC}(\theta|\y_\text{obs})$ is also an approximation to the true posterior, the bias being controlled by the bandwidth $h$, as shown in Figure \ref{fig_abc_noisy_lik}(d). 
\newline
\newline
After selecting a suitable $\epsilon$ and $M$, as discussed above, the Monte Carlo approximation to $m_\text{ABC}(\theta | \y_\text{obs})$ can be computed via noisy Monte Carlo algorithms. A surrogate of $\ell_\text{ABC}(\theta)$ can be built from a set $\{\theta_{j},\widetilde{\ell}_{\text{ABC}}(\theta_j)\}_{j=1}^J$ and used for acceleration. Alternatively, one can build a surrogate of the discrepancy function $\Delta_{\theta}=\Delta(\theta)$ via $\{\theta_{j},\Delta(\theta_j)\}_{j=1}^J$, which is a random function due to $\y_{\theta}$ being random. The advantages of modeling $\Delta_{\theta}$ instead of $\ell_\text{ABC}(\theta)$ are discussed in \cite{Gutmann15,jarvenpaa2019efficient}, for instance, $\Delta_{\theta}$ is independent of the bandwidth choice (and, more generally, of the kernel $h(\y_\text{obs}|\y,\epsilon,\theta)$). Furthermore, for some kernels, including $\mathbbm{1}\left( \Delta_{\theta} < \epsilon \right)$,  minimizing $\Delta_{\theta}$ corresponds to maximizing a lower bound of $\ell_\text{ABC}(\theta)$ \cite{Gutmann15}. 
Hence, a Gaussian process (GP) surrogate of $\Delta_{\theta}$ can be used in conjunction with Bayesian optimization (BO) or active learning algorithms to efficiently design the set $\{\theta_{j},\Delta(\theta_j)\}_{j=1}^J$ \cite{Gutmann15,jarvenpaa2019efficient}. 
In Figure \ref{fig_discrepancy}(a), we show the stochastic nature of $\Delta_{\theta}$, which we model with a GP, as shown in Figure \ref{fig_discrepancy}(b). The set of $J=12$ design points $\{\theta_{j},\Delta(\theta_j)\}_{j=1}^J$ have been obtained by first sampling 2 points from the prior and running a BO for 10 iterations. At each iteration, the next point to be evaluated is selected by minimizing the so-called lower confidence bound (LCB), 
$$
A_t(\theta) = \mu_{GP}^{(t)}(\theta) - \beta_t \sigma^{(t)}_{GP}(\theta),
$$
where $\mu_{GP}^{(t)}(\theta)$ and $\sigma^{(t)}_{GP}(\theta)$ are the posterior moments of the GP at iteration $t$ and $\beta_t$ is a parameter that tradeoffs exploration with exploitation. 
Finally, in Figure \ref{fig_discrepancy}(c), we show the  surrogate $\widehat{\ell}_\text{ABC}(\theta)$ that resulted from $\widehat{\Delta}(\theta)$.
\newline
\newline
We run two experiments and compare the relative square error in estimating the moments of the ABC posterior, namely the mean $\mu_{\text{abc}}$ and variance $\sigma^2_{\text{abc}}$ of four different methods. We compare:
\begin{itemize}
\item[{\bf A1}] the (naive) standard accept-reject ABC, which corresponds to noisy IS using the prior as proposal pdf;
 \item[{\bf A2}]  a (non-noisy) IS where we use a fixed GP surrogate of the ABC likelihood built with GP, shown in Fig. \ref{fig_discrepancy}(c), as the true likelihood;
 \item[{\bf A3}]  a non-iterative version of noisy deep IS (N-DIS) where the surrogate is used as proposal by using resampling steps \cite{llorente2021deep}, that in case coincides with sampling-importance-resampling (SIR);
  \item[{\bf A4}] finally, we consider another version of N-DIS where the surrogate is directly built using the noisy likelihood evaluations with k-nearest neighbors (k-NN) with $K=10$ and in an online fashion starting with samples from the prior.
\end{itemize}
  For a fair comparison, in all the methods, we keep the same number of noisy evaluations $E=1000$ (always using $M=1$), except for {\bf A2}, that targets a fixed surrogate built offline, hence it does not require obtaining noisy evaluations. Our goal is to analyze the potential performance gains of using surrogates, while we increase the ``conflict'' between the prior and the likelihood. 
\newline
\newline
First, we fixed the bandwidth to $\epsilon=0.1$ and test different values of the prior standard deviation, $\sigma_0 \in \{1,2,3,5\}$. A prior with greater variance is beneficial to all the methods, but especially to the simple accept-reject ABC algorithm. We can see  in Figure \ref{fig_error_abc}(a)-(b) that, except for algorithm {\bf A2} that has a large bias, the algorithms have similar performance when increasing the prior width. When the prior is very informative, namely $\sigma_0=1,2$, algorithm {\bf A2} improves the results of the baseline. For those values, the use of surrogate accelerates the finding of the region of high probability.  As a second experiment, we fixed $\sigma_0=3$ and test the values of the bandwidth $\epsilon\in\{0.1,1,5,10\}$. The conclusions are also that, only when the bandwidth is small (and hence there is more mismatch between prior and likelihood), the algorithm {\bf A2} shows better performance than the baseline. 
We note that algorithm {\bf A3} is not able to beat the baseline, probably because of two reasons. It does not use a clever initialization for the surrogate, contrary to algorithm {\bf A3}  that builds the surrogate of the likelihood indirectly with Bayesian optimization. Moreover, the noisy evaluations are frequently 0's, especially far from the true value of $\theta$ used to generate the data, causing the k-NN surrogate to initially rule out important regions of the parameter space.

\begin{figure}[h!]
	\centering
	\begin{subfigure}[b]{0.35\textwidth}
		\includegraphics[width=1\textwidth]{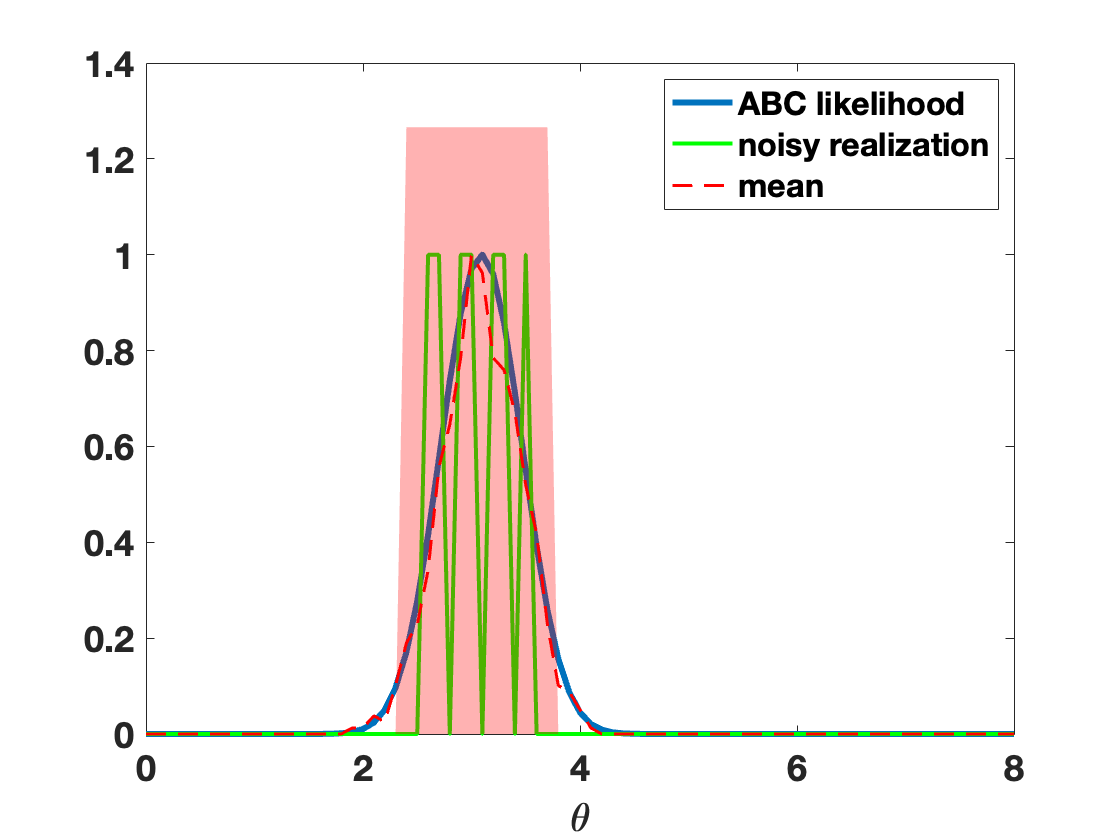}
			\caption{$M=1$}
	\end{subfigure}
	\begin{subfigure}[b]{0.35\textwidth}
		\includegraphics[width=1\textwidth]{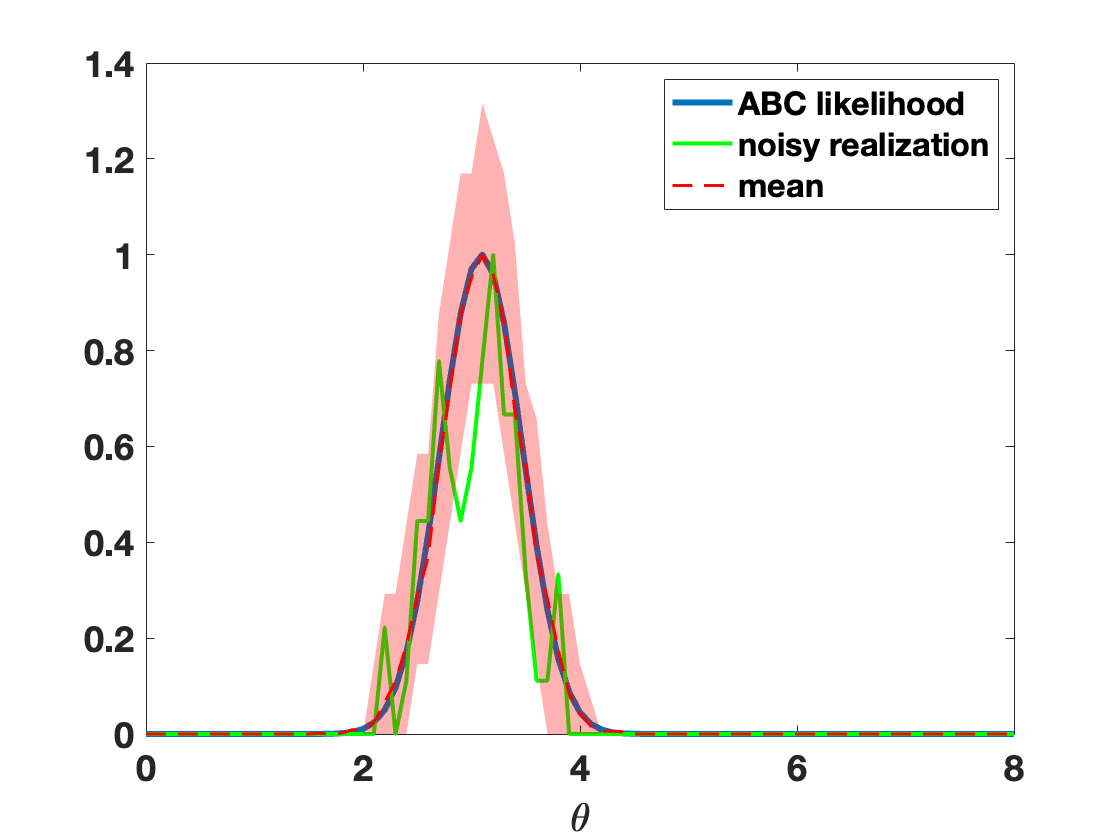}
				\caption{$M=10$}
	\end{subfigure}
	
	\begin{subfigure}[b]{0.35\textwidth}
		\includegraphics[width=1\textwidth]{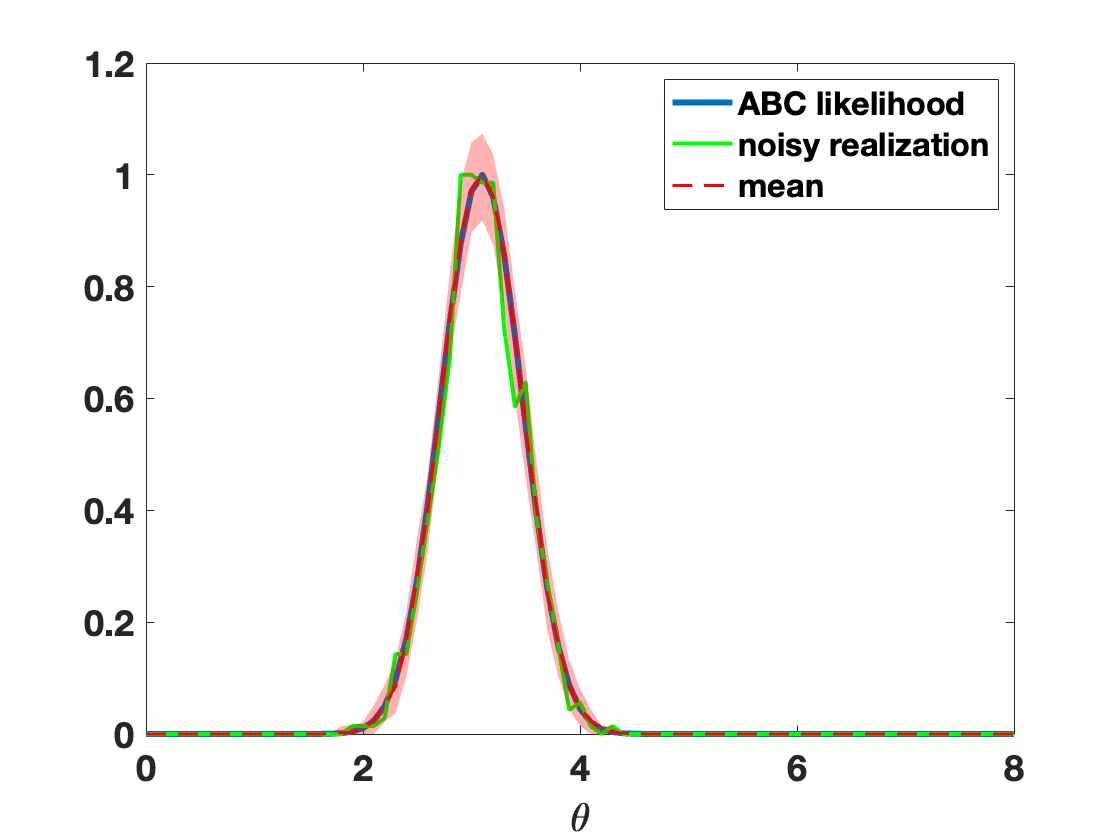}
				\caption{$M=100$}
	\end{subfigure}
	\begin{subfigure}[b]{0.35\textwidth}
	\includegraphics[width=1\textwidth]{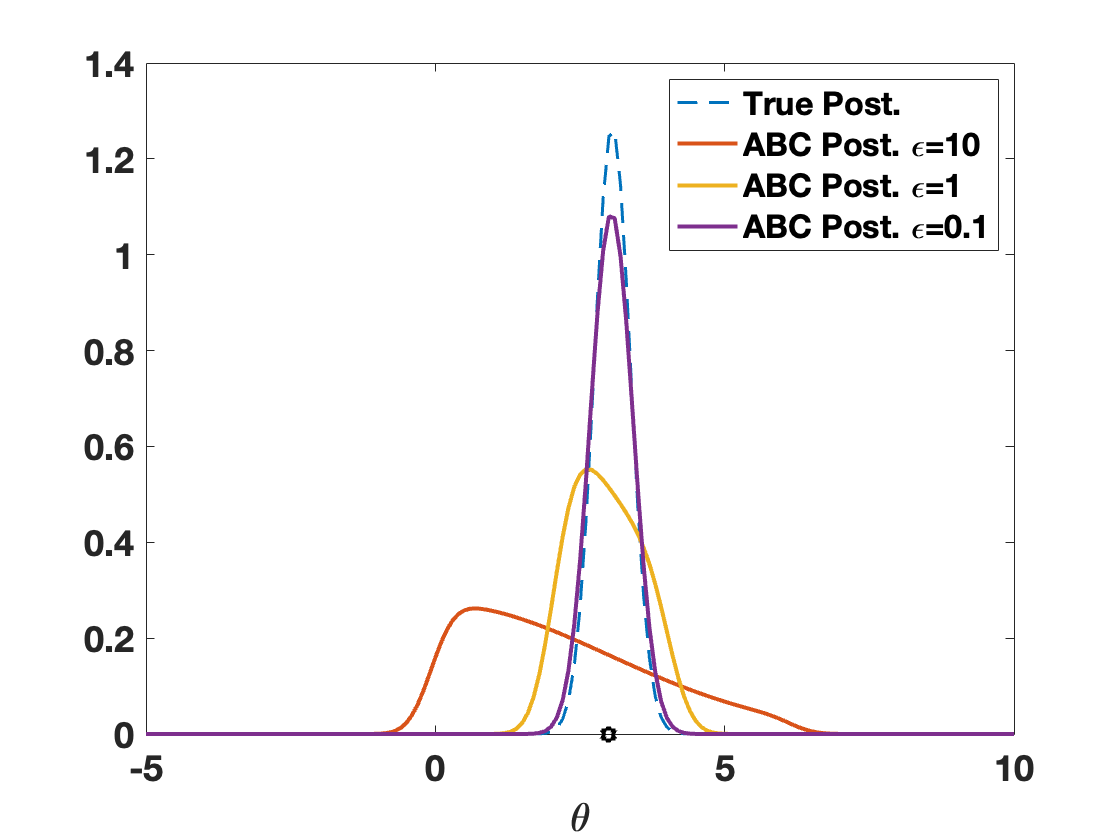}
			\caption{True and ABC posteriors}
\end{subfigure}
\
	\caption{\label{fig_abc_noisy_lik} 
	{
	Figures (a)-(b)-(c) show noisy realizations of the ABC likelihood with bandwidth $\epsilon=0.1$ for $M\in \{1,10,100\}$, respectively. We also plot the 0.1 and 0.9 quantiles of the noisy realizations.	Figure (d) shows the true posterior distribution employing the Gaussian likelihood along with three ABC posteriors with bandwidths $\epsilon\in\{0.1,10,100\}$. The true value used to generate the observed data is also depicted.
	}
		}
\end{figure}

\begin{figure}[h!]
	\centering
	\begin{subfigure}[b]{0.3\textwidth}
		\includegraphics[width=1\textwidth]{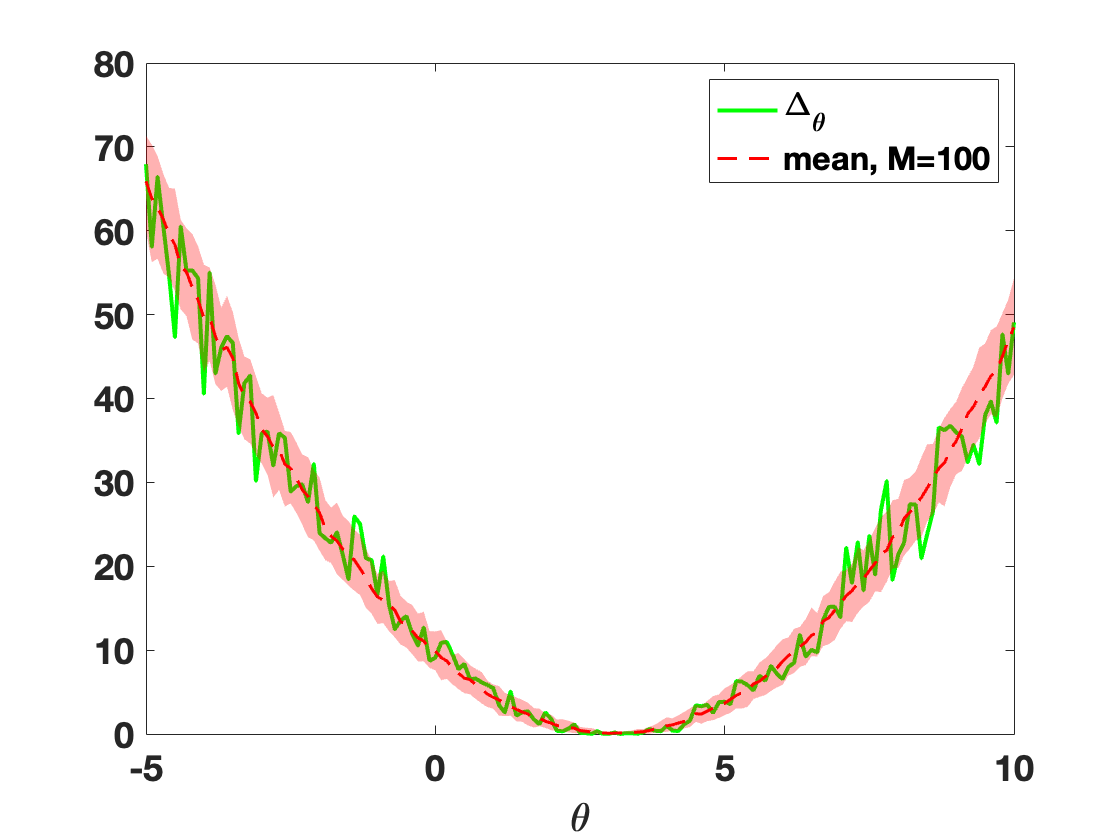}
			\caption{}
	\end{subfigure}
	\begin{subfigure}[b]{0.3\textwidth}
		\includegraphics[width=1\textwidth]{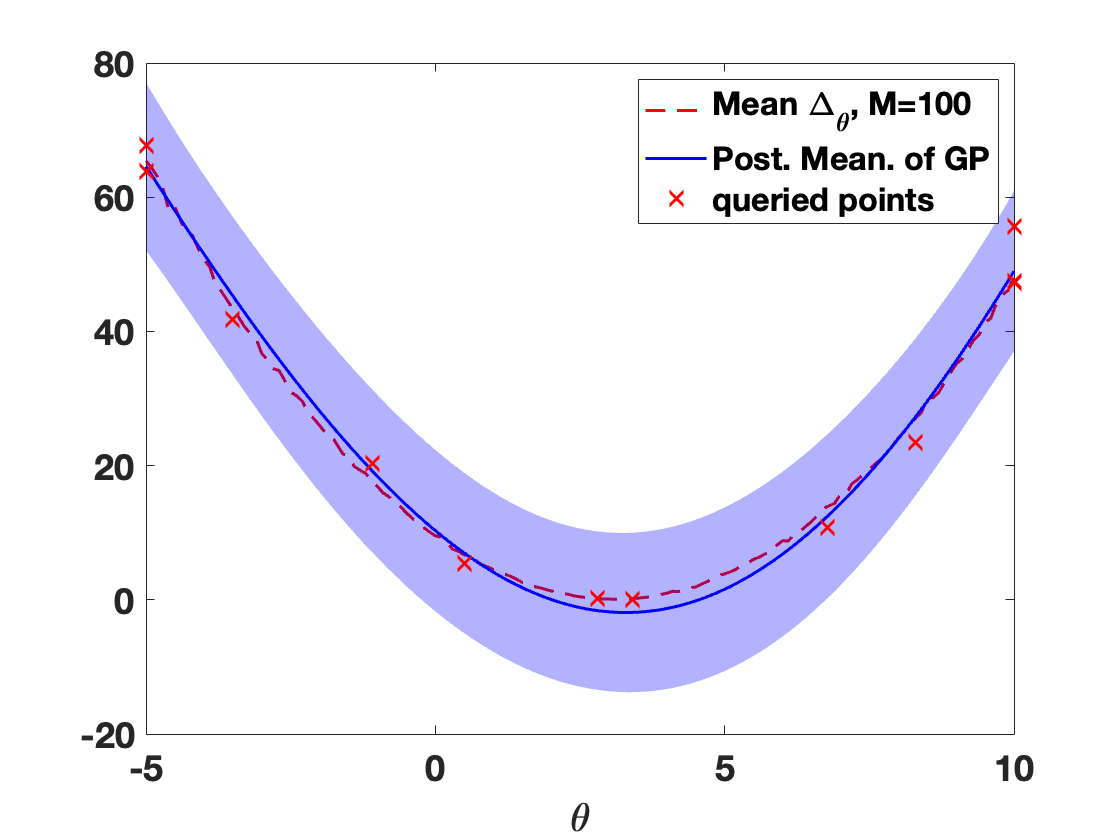}
				\caption{}
	\end{subfigure}	
	\begin{subfigure}[b]{0.3\textwidth}
		\includegraphics[width=1\textwidth]{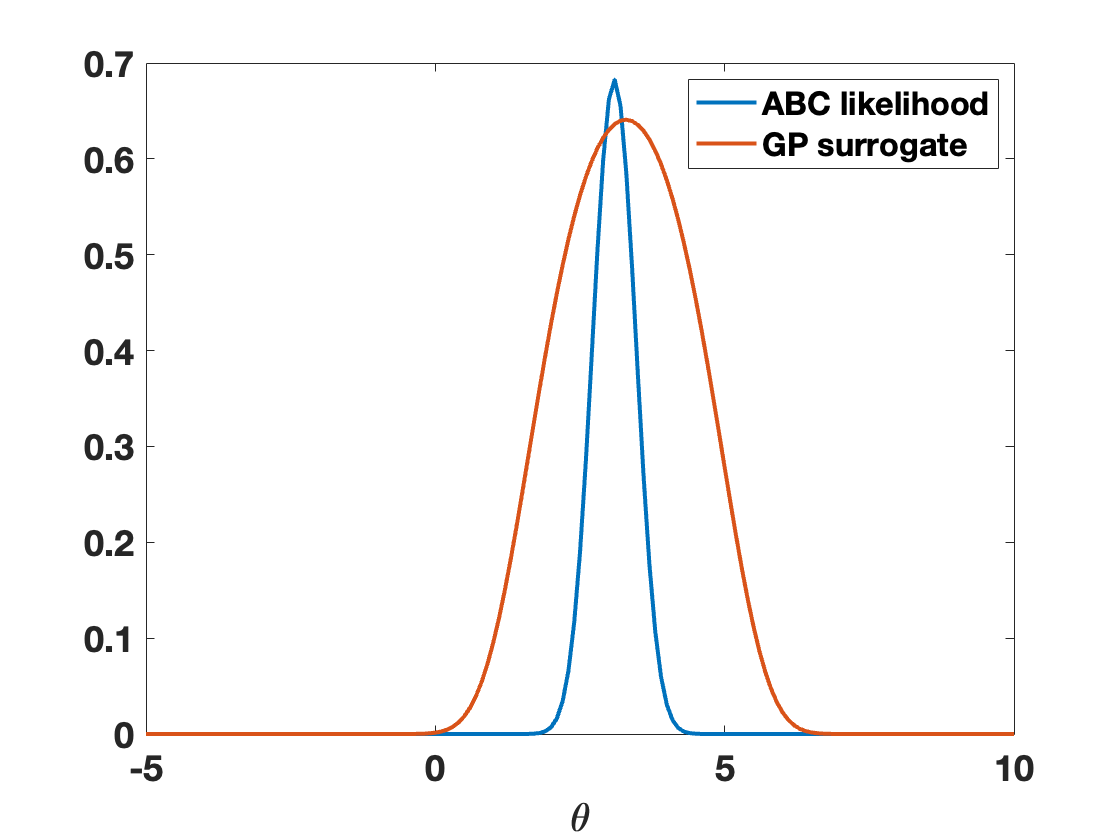}
				\caption{}
	\end{subfigure}
	\
	\caption{\label{fig_discrepancy} 
	{ Figure (a) shows a noisy realization of the discrepancy function $\Delta_{\theta}$, its mean and the 0.1 and 0.9 quantiles computed with $M=100$ for each $\theta$. Figure (b) shows the posterior mean and quantiles of the GP surrogate of $\Delta_{\theta}$, where the nodes have been sequentially selected in order to balance exploration and exploitation. Figure (c) shows the resulting surrogate of the ABC likelihood obtained from the surrogate of $\Delta_\theta$.
		}
		}
\end{figure}

\begin{figure}[h!]
	\centering
	\begin{subfigure}[b]{0.35\textwidth}
		\includegraphics[width=1\textwidth]{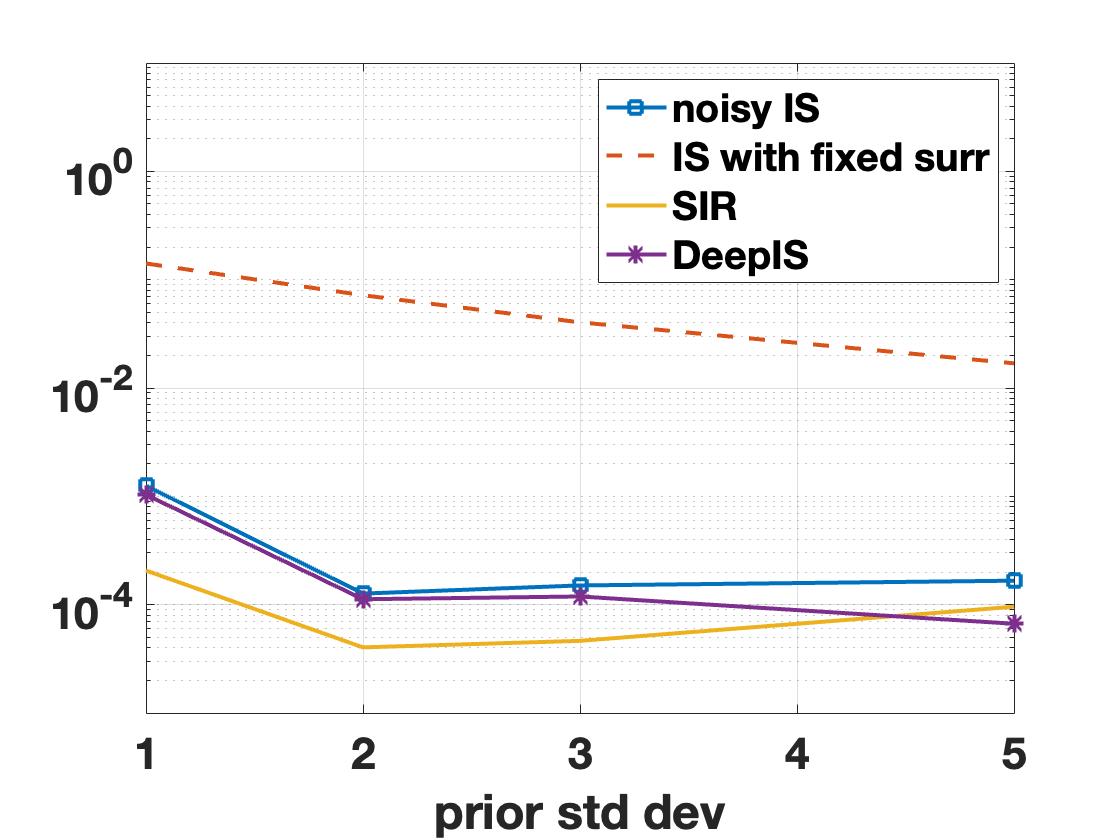}
			\caption{$\mu_{\text{abc}}$}
	\end{subfigure}
	\begin{subfigure}[b]{0.35\textwidth}
		\includegraphics[width=1\textwidth]{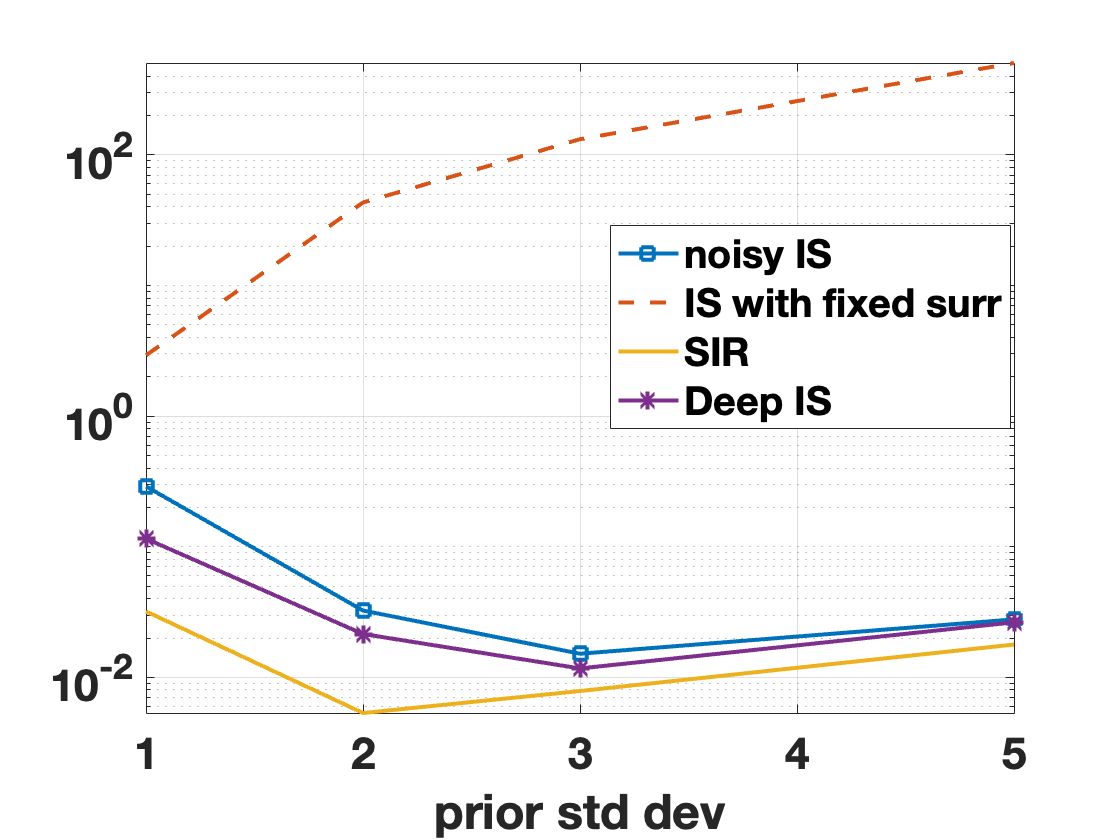}
				\caption{$\sigma^2_{\text{abc}}$}
	\end{subfigure}	
	
	\begin{subfigure}[b]{0.35\textwidth}
		\includegraphics[width=1\textwidth]{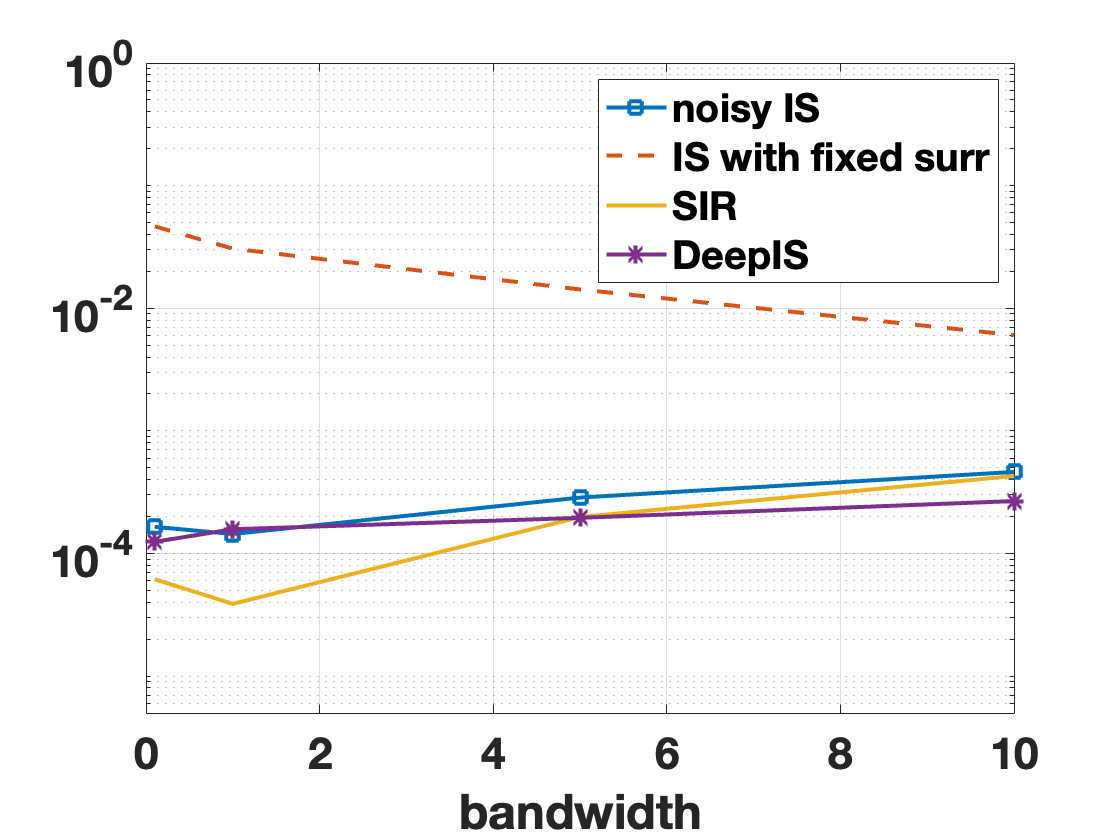}
			\caption{$\mu_{\text{abc}}$}
	\end{subfigure}
	\begin{subfigure}[b]{0.35\textwidth}
		\includegraphics[width=1\textwidth]{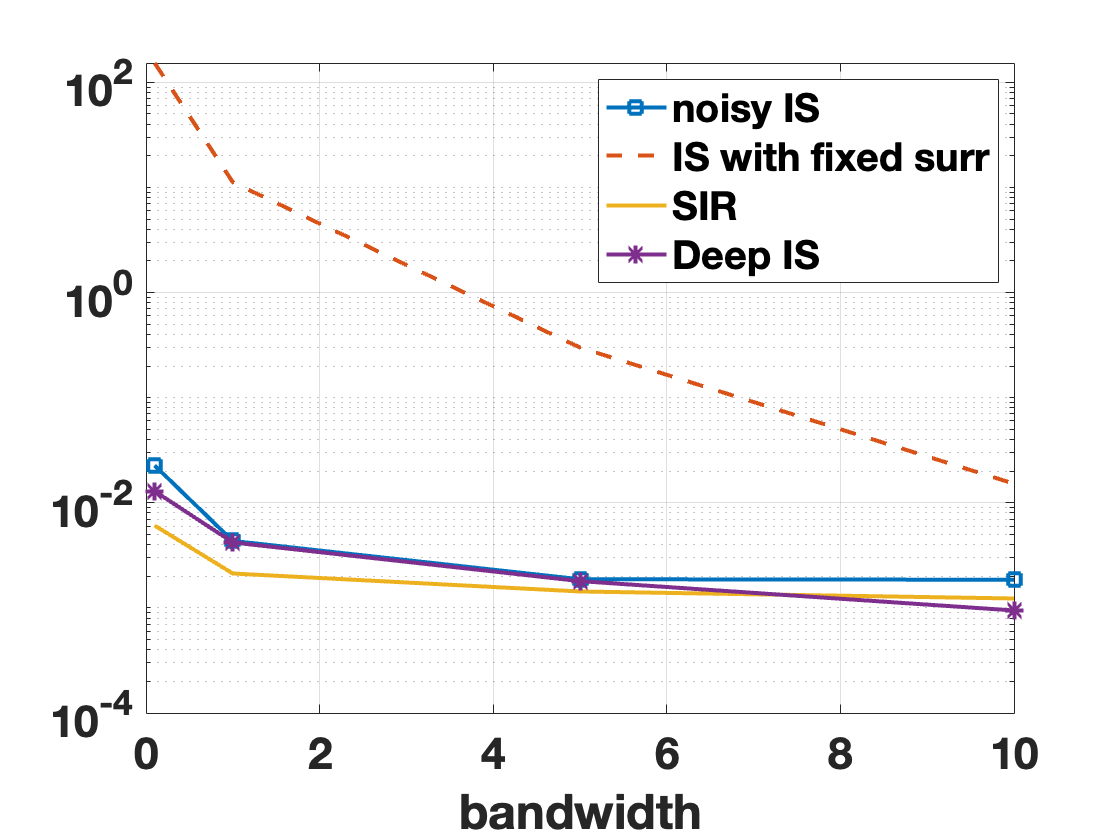}
				\caption{$\sigma^2_{\text{abc}}$}
	\end{subfigure}	
	\
	\caption{\label{fig_error_abc} 
	{ Figures (a)-(b) show respectively the relative square error in estimating the posterior mean and variance of the ABC posterior versus the standard deviation of the prior, with fixed bandwidth. Figures (c)-(d) show respectively the relative square error in estimating the posterior mean and variance of the ABC posterior versus the bandwidth, with fixed prior standard deviation.
		}
		}
\end{figure}

}

\section{{Summary and} conclusions}\label{sec_conc}

We have provided an overview of Monte Carlo methods which use surrogate models built with regression techniques, for dealing with noisy and costly densities. Indeed, by employing surrogate models, we can avoid the evaluation of expensive true models and perform a smoothing of the noisy realizations. This has important implications for performance in real-world applications. 
\newline
We have described a general joint framework which encompasses most of the techniques in the literature. We have given a classification of the analyzed techniques in three main families. We have highlighted the connections and differences among the algorithms by means of  several explanatory tables and figures. The range of application of the methods have been discussed. Specifically,   a detailed description of the likelihood-free approach and the reinforcement learning setting is presented.
\newline
Numerical simulations have shown that, generally,  the use of surrogates can improve the performance of the algorithms. Indeed, the surrogate plays the role of an adaptive non-parametric proposal which is adapted using not only the spatial information contained in the samples, but also the noisy evaluations of the target. This increases the efficiency of the corresponding Monte Carlo estimators since it fosters the exploration of the space. 
On the other hand, pathological constructions of the surrogate, i.e., when the surrogate takes small values in high probability regions of the target pdf, can jeopardize the performance of the algorithms, at least in the first iterations. Furthermore, the correction step in the exact algorithms yields more robust schemes.
\newline
\newline
{
The design of a specific scheme depends { {\it conceptually}} on the following main elements:
\begin{itemize}
\item MCMC or IS approach;
\item choose of the surrogate regression method;
\item select one of the three classes, C1, C2 or C3;
\item decide the acquisition procedure for adding new nodes for improving the surrogate function.
\end{itemize}
 The choice of one particular scheme depends on the specific application and computational cost that each elements have in that particular framework. For instance, if obtaining the noisy realizations $\widetilde{p}_M$ is cheap perhaps the use of a surrogate can be completely avoided, applying the vanilla scheme in Section \ref{VanillaSchemesSect} and spend more computational effort in reducing bias and variance increasing $M$ (if needed).  If prior information is available and the construction of the surrogate can be fostered by a good initialization, the second class C2 of methods is probably the most adequate (avoiding to waste computational time with additional correction steps). If no prior information is available, and the user could not ``trust'' the construction of the surrogate in the first iterations and, at the same time, desires to bound the number of nodes $J$, the third class C3 of methods is probably more appropriate. Indeed, the additional correction step ensures a proper sampling even from the first iteration and can also help in incorporating suitable nodes (in regions where more nodes are required). If acquiring noisy realizations $\widetilde{p}_M$  if extremely costly, the construction of a proper surrogate is a fundamental tool and, for this purpose, the use of smart acquisition function is crucial in order to keep the evaluation cost of the (non-parametric) surrogate parsimonious (i.e., keeping $J$ small). { Clearly, any 
 parallelization strategies proposed in the literature can be applied also in this framework,
 helping reducing even more cost and/or possibly fostering a better construction of the surrogate model. 
 }
 \newline
Finally it deserves to be mentioned that, for all the classes, it is important to avoid the overfitting in the construction of the surrogate in the first iterations, since this fosters the exploration of the space. Future works should be focused on a depth study addressing only one the each of the main elements listed above, that form the algorithm structures.  Indeed, each elements is formed by numerous several options that require a specific analysis and comparison.
 }




\bibliographystyle{plainnat}
\bibliography{bibliografia}

\appendix

\section{Proof for noisy MH algorithm}\label{app_noisyMH}

We provide here a simple proof showing that the invariant density of a MH algorithm using noisy realizations $\widetilde{p}_M( \vtheta)$ is $m(\vtheta)$ (i.e. a pseudo-marginal MH algorithm). For more details see \cite{andrieu2009pseudo,andrieu2015convergence}.
Let us consider the acceptance ratio of the noisy MH algorithm
\begin{align}
r(\vtheta_{t-1},\vtheta_\text{prop}) = \frac{\widetilde{p}_M( \vtheta_\text{prop}) \varphi(\vtheta_{t-1}|\vtheta_\text{prop})}{\widetilde{p}_M( \vtheta_{t-1}) \varphi(\vtheta_\text{prop}|\vtheta_{t-1})}.
\end{align}
Now, let us rewrite it as
\begin{align}
r(\vtheta_{t-1},\vtheta_\text{prop}) = \frac{
	\frac{\widetilde{p}_M( \vtheta_\text{prop})}{m(\vtheta_\text{prop})}m(\vtheta_\text{prop})	
	\varphi(\vtheta_{t-1}|\vtheta_\text{prop})
}
{
	\frac{\widetilde{p}_M( \vtheta_{t-1})}{m(\vtheta_{t-1})}m(\vtheta_{t-1})
	\varphi(\vtheta_\text{prop}|\vtheta_{t-1})
}.
\end{align}
Define $\lambda = \frac{\widetilde{p}_M( \vtheta)}{m(\vtheta)}$ as a random variable with pdf given by $g(\lambda|\vtheta)$. Note that $\mathbb{E}[\lambda|\vtheta] \propto 1$ for any $\vtheta$. Denoting $\lambda_\text{prop} = \frac{\widetilde{p}_M( \vtheta_\text{prop})}{m(\vtheta_\text{prop})}$ and $\lambda_{t-1} = \frac{\widetilde{p}_M( \vtheta_{t-1})}{m(\vtheta_{t-1})}$, multiplying by $g(\lambda_\text{prop}|\vtheta_\text{prop})g(\lambda_{t-1}|\vtheta_{t-1})$ in both numerator and denominator, and rearranging the terms we see that the acceptance ratio is
\begin{align}
r(\vtheta_{t-1},\vtheta_\text{prop}) = \frac{
	\lambda_\text{prop}m(\vtheta_\text{prop})	
	g(\lambda_\text{prop}|\vtheta_\text{prop})
	\varphi(\vtheta_{t-1}|\vtheta_\text{prop})
	g(\lambda_{t-1}|\vtheta_{t-1})
}
{
	\lambda_{t-1}m(\vtheta_{t-1})
	g(\lambda_{t-1}|\vtheta_{t-1})
	\varphi(\vtheta_\text{prop}|\vtheta_{t-1})
	g(\lambda_\text{prop}|\vtheta_\text{prop})
}.
\end{align}
Now, let us define $q_\text{equiv}(\vtheta,\lambda|\vtheta',\lambda') = g(\lambda|\vtheta)\varphi(\vtheta|\vtheta')$ as the equivalent proposal in the joint space $(\vtheta,\lambda)$. 
Hence, the ratio is finally expressed as
\begin{align}
r(\vtheta_{t-1},\vtheta_\text{prop},\lambda_{t-1},\lambda_\text{prop}) = \frac{
	\lambda_\text{prop}m(\vtheta_\text{prop})	
	g(\lambda_\text{prop}|\vtheta_\text{prop})
	q_\text{equiv}(\vtheta_{t-1},w_{t-1}|\vtheta_\text{prop},\lambda_\text{prop})
}
{
	\lambda_{t-1}m(\vtheta_{t-1})
	g(\lambda_{t-1}|\vtheta_{t-1})
	q_\text{equiv}(\vtheta_\text{prop},w_\text{prop}|\vtheta_{t-1},\lambda_{t-1})
}.
\end{align}
It can be seen now that the invariant density is proportional to $\lambda\cdot m(\vtheta)\cdot g(\lambda|\vtheta)$, whose marginal is $\int \lambda\ m(\vtheta)g(\lambda|\vtheta)d\lambda \propto m(\vtheta)$.

\section{Proof for noisy IS}\label{App_noisyIS}

We show that an IS estimator built with noisy realizations $\widetilde{p}_M( \vtheta)$, converges to expectations w.r.t.\ $m(\vtheta)$.
Let $q(\vtheta)$ denote a proposal pdf, and let 
\begin{align}
\widetilde{Z} = \frac{1}{N}\sum_{i=1}^{N}\frac{\widetilde{p}_M( \vtheta_i)}{q(\vtheta_i)}=\frac{1}{N}\sum_{i=1}^{N}\widetilde{w}_i,
\end{align}
be the IS estimator built with noisy realizations, where $\widetilde{w}_i=\frac{\widetilde{p}_M( \vtheta_i)}{q(\vtheta_i)}$ are the noisy weights, and $\{\vtheta_i\}_{i=1}^N$ are iid samples from $q$.
The non-noisy IS estimator, recalling $m(\vtheta)=p(\vtheta)-\mu_M(\vtheta)$,
\begin{align}
\widehat{Z} = \frac{1}{N}\sum_{i=1}^{N}\frac{m(\vtheta_i)}{q(\vtheta_i)}=\frac{1}{N}\sum_{i=1}^{N}{w}_i,
\end{align} 
is an unbiased estimator of $Z = \int m(\vtheta)d\vtheta$, i.e., $\Exp[\widehat{Z}]=Z$, converging as $N \to \infty$ at rate $\frac{1}{N}$.
We aim to show that $\widetilde{Z}$ is also an unbiased estimator of $Z$, with greater variance than $\widehat{Z}$, but the same convergence speed, i.e., its variance decreases at $\frac{1}{N}$ rate. 

Let $\bm{\Theta}=(\vtheta_1,\dots,\vtheta_N)$ denote the $N$ samples from $q$.
By the law of total expectation, we have that $\Exp[\widetilde{Z}] = \Exp\left[\Exp[\widetilde{Z}|\bm{\Theta}]\right]$. In the inner expectation, we use the fact the $\widetilde{w}_i$'s are i.i.d., hence
\begin{align}
\Exp[\widetilde{Z}|\bm{\Theta}] = \frac{1}{N}\sum_{i=1}^{N}\Exp[\widetilde{w}_i|\vtheta_i] 
= \frac{1}{N}\sum_{i=1}^{N}\frac{1}{q(\vtheta_i)}\Exp[\widetilde{p}_M( \vtheta_i)|\vtheta_i] = \widehat{Z},
\end{align}
and
\begin{align}
\Exp[\widetilde{Z}] = \Exp\left[\Exp[\widetilde{Z}|\bm{\Theta}]\right] = \Exp[\widehat{Z}] = Z.
\end{align}
By the law of total variance, we have that
\begin{align}
\texttt{var}[\widetilde{Z}] = \Exp\left[\texttt{var}[\widetilde{Z}|\bm{\Theta}]\right] + \texttt{var}\left[\Exp[\widetilde{Z}|\bm{\Theta}]\right].
\end{align}
Using the above result, we have that the second term is
\begin{align}
\texttt{var}\left[\Exp[\widetilde{Z}|\bm{\Theta}]\right] = \texttt{var}[\widehat{Z}] = \mathcal{O}\left(1/N\right).
\end{align}
Regarding the first term, we have
\begin{align}
\texttt{var}[\widetilde{Z}|\bm{\Theta}] &=\frac{1}{N^2}\sum_{i=1}^{N}\texttt{var}[\widetilde{w}_i|\vtheta_i] =\frac{1}{N^2}\sum_{i=1}^{N}\frac{1}{q(\vtheta_i)^2}\texttt{var}[\widetilde{p}_M( \vtheta_i)|\vtheta_i] \nonumber\\
&=\frac{1}{N^2}\sum_{i=1}^{N}\frac{s^2(\vtheta_i)}{q(\vtheta_i)^2}.
\end{align}
Assuming that $\frac{s^2(\vtheta)}{q(\vtheta)}<\infty$ for all $\vtheta$, we have that
\begin{align}
\Exp\left[\texttt{var}[\widetilde{Z}|\bm{\Theta}]\right] 
&=\frac{1}{N^2}\sum_{i=1}^{N}\Exp\left[\frac{s^2(\vtheta_i)}{q(\vtheta_i)^2}\right]
=\frac{1}{N}\Exp\left[\frac{s^2(\vtheta)}{q(\vtheta)^2}\right].
\end{align}
Hence, we finally have that
\begin{align}
\texttt{var}[\widetilde{Z}] &= \frac{1}{N}\Exp\left[\frac{s^2(\vtheta)}{q(\vtheta)^2}\right] + \texttt{var}[\widehat{Z}] = \mathcal{O}\left(\frac{1}{N}\right) \nonumber \\ 
&\geq \texttt{var}[\widehat{Z}].
\end{align}
From this expression, we can deduce that the variance of $\widetilde{Z}$ depends on the mismatch between $q(\vtheta)$ and $s^2(\vtheta)$.
Proving that the noisy IS estimator $\widetilde{I} = \frac{1}{N}\sum_{i=1}^{N}\frac{\widetilde{p}_M( \vtheta)f(\vtheta)}{q(\vtheta)}$ converges to $I = \int f(\vtheta)m(\vtheta)d\vtheta$ is immediate. 
Thus, the ratio 
$\frac{\widetilde{I}}{\widetilde{Z}} = \frac{1}{\sum_{j=1}^N\widetilde{w}_j}\sum_{i=1}^N\widetilde{w}_i f(\vtheta_i),$
(i.e. the noisy self-normalized IS estimator) is a consistent estimator of $\frac{\int_{{\bm \Theta}} f(\vtheta)m(\vtheta)d\vtheta}{\int_{{\bm \Theta}} m(\vtheta)d\vtheta}$.

\section{ilustrative example}\label{illustrative_ex_1D}
\begin{figure}[h!]
	\centering
	\begin{subfigure}[b]{0.31\textwidth}
		\includegraphics[width=1\textwidth]{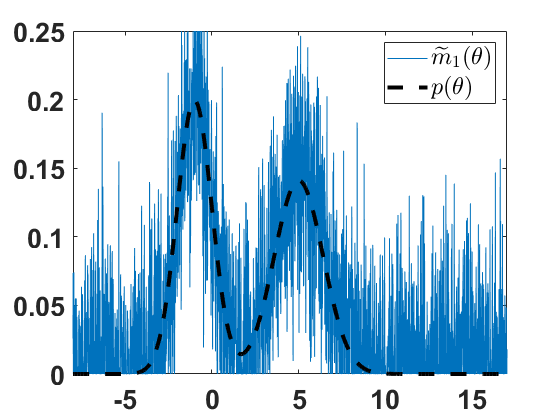}
		\caption{$\widetilde{p}_1(\theta)$}
	\end{subfigure}
	\begin{subfigure}[b]{0.31\textwidth}
		\includegraphics[width=1\textwidth]{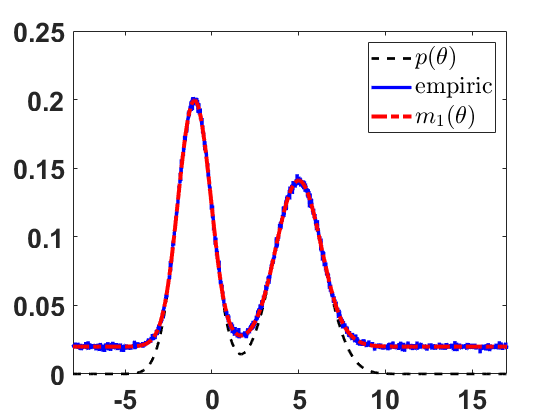}
		\caption{$m_1(\theta)$ and $p(\vtheta)$}
	\end{subfigure}
	\begin{subfigure}[b]{0.31\textwidth}
	\includegraphics[width=1\textwidth]{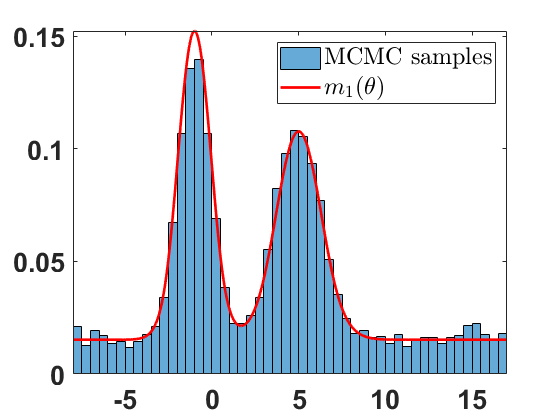}
	\caption{histogram of samples}
	\end{subfigure}
	\caption{\label{fig_target_ruidosa_1D}
		{(\bf a)} The target pdf $p(\theta)$ and  a realization of $\widetilde{p}_1(\theta)=\max(0,p(\theta)+\epsilon)$. {(\bf b)} Again  the target pdf $p(\theta)$ (dashed line), the  mean function $m(\theta)=\mathbb{E}[\widetilde{p}_M( \theta)]$ and its empirical approximation averaging several realizations. {(\bf c)}  Histogram of the samples generated by a noisy MCMC scheme.  }
\end{figure}

\begin{figure}[h!]
	\centering
	\begin{subfigure}[b]{0.35\textwidth}
		\includegraphics[width=1\textwidth]{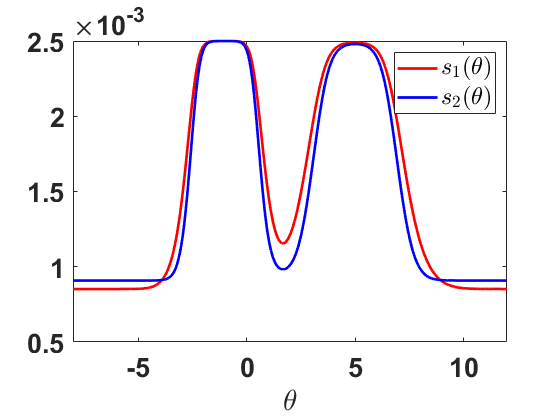}
		\caption{$s_i(\theta)$ vs $\theta$}
	\end{subfigure}
	\begin{subfigure}[b]{0.35\textwidth}
	\includegraphics[width=1\textwidth]{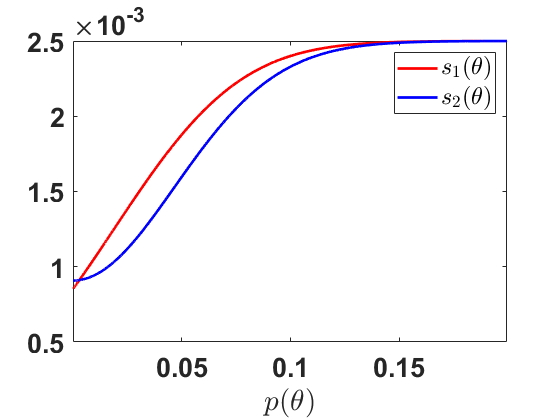}
	\caption{$s_i(\theta)$ vs $p(\theta)$}
    \end{subfigure}
	\caption{\label{fig:toy1D}
	Behavior of the variance $s_i^2(\theta)$ in both models.	
		{\bf (a)} On the left: Plots of  $\sqrt{s_i^2(\theta)}$ versus $\theta$ for $i=1,2$.
		{\bf (b)} Plots of $\sqrt{s_i^2(\theta)}$ versus $p(\theta)$  for $i=1,2$..}
\end{figure}

As an illustration, let us consider the one-dimensional density $p(\theta) = \frac{1}{2}\mathcal{N}(\theta;-1,1)+
\frac{1}{2}\mathcal{N}(\theta;5,2),$ restricted in the finite domain $[-8,17]$,
and two noisy versions
$$
\widetilde{p}_1(\theta) = \max(0,p(\theta)+\epsilon),\ \text{and}\ \widetilde{p}_2(\theta) = |p(\theta)+\epsilon|,
$$
where $\epsilon \sim \mathcal{N}(0,0.05^2)$. Namely, $\widetilde{p}_i(\theta)$, $i=1,2$, correspond to rectified Gaussian and folded Gaussian random variables, respectively (for any $\theta$). In Figure \ref{fig_target_ruidosa_1D}-(a), we show one realization of $\widetilde{p}_1(\theta)$. 
In Figure \ref{fig_target_ruidosa_1D}-(b), we show the average of $\widetilde{p}_1(\theta)$ (empirically and theoretically).
In Figure \ref{fig_target_ruidosa_1D}-(c), we show the histogram of samples obtained by running a (pseudo-marginal) MH algorithm on $\widetilde{p}_1(\vtheta)$. 
In these cases, the expected values do not coincide with $p(\theta)$, i.e., $m_i(\theta)\neq p(\theta)$. Analytical expressions of $m_i(\theta)$, as well as $s_i^2(\theta)$, can be obtained as shown in App.\ \ref{App_noise_models}.  The variance behaviors are depicted in Figures \ref{fig:toy1D}(a)--(b).

\subsection{Analytical expressions of the noise models in illustrative example} \label{App_noise_models}

Let $\epsilon \sim \mathcal{N}(0,\sigma^2)$.
The analytical expressions of $m(\vtheta)$ for the noise models in the illustrative example of Sect. \ref{GenFrame} are provided here.
\newline
{\bf Rectified Gaussian.} By setting $\widetilde{p}_M( \vtheta) = \max(0,p(\vtheta)+\epsilon)$, then $\widetilde{p}_M( \vtheta)|\vtheta \sim \mathcal{N}^\text{R}(p(\vtheta),\sigma^2)$ is a rectified Gaussian random variable, whose mean is
$$
m(\vtheta) = \left[p(\vtheta) + \sigma\frac{\phi(-p(\vtheta)/\sigma)}{1-\bm{\Phi}(-p(\vtheta)/\sigma)}\right]\left[1-\bm{\Phi}(-p(\vtheta)/\sigma)\right],
$$
where $\phi(\theta)$ and $\bm{\Phi}(\theta)$ are the pdf and cdf, respectively, of the standard normal distribution.
\newline
\newline		
{\bf Folded Gaussian.} The random variable $\widetilde{p}_M( \vtheta) = |p(\vtheta)+\epsilon|$ corresponds to a folded Gaussian random variable.
We have
$$
m(\vtheta) = \sigma\sqrt{\frac{2}{\pi}}\exp\left(-p^2(\vtheta)/2\sigma^2\right) + p(\vtheta)[1-2\bm{\Phi}(-p(\vtheta)/\sigma)].
$$

{

\section{Emulation of the log-likelihood with a Gaussian process}\label{EmuLogLikeSect}

Let us consider the scenario when the likelihood function $\ell(\y|\vtheta)$ is intractable, and we use Gaussian processes (GPs) to emulate the log-likelihood using noisy observations. In this setting, the function $L(\vtheta) = \log\ell(\y|\vtheta)$ is assumed a sample from a GP prior $\mathcal{GP}(0,k_{\vpsi}(\vtheta,\vtheta'))$, where $k_{\vpsi}(\vtheta,\vtheta')$ is the covariance function with hyperparameters $\vpsi$, encoding the smoothness assumptions about $L(\vtheta)$ \cite{rasmussen06,MARTINO202117}. The noisy observations are assumed to be Gaussian distributed, $\widetilde{L}(\vtheta_j) \sim \mathcal{N}(L(\vtheta_j),\sigma_{\text{noise}}^2)$.
Conditional on the observations $\mathcal{D} = \{\vtheta_i,\widetilde{L}(\vtheta_j)\}_{i=1}^J$, the posterior of $L(\vtheta)$ is also a GP,
\begin{align*}
L(\vtheta)|\mathcal{D} &\sim \mathcal{GP}(\mu_p(\vtheta),k_p(\vtheta,\vtheta')), \\
\mu_p(\vtheta) &= {\bf k}_J^\top(\vtheta)(K + \sigma^2_{\text{noise}})^{-1}\widetilde{{\bf L}},\\
k_p(\vtheta,\vtheta') &= k_{\vpsi}(\vtheta,\vtheta') - {\bf k}_J^\top(\vtheta)({\bf K} + \sigma^2_{\text{noise}})^{-1}{\bf k}_J(\vtheta'), \quad
\sigma^2_p(\vtheta) = k_p(\vtheta,\vtheta),
\end{align*}
where  $\widetilde{\bf L} = [\widetilde{L}(\vtheta_1),\dots,\widetilde{L}(\vtheta_J)]$, ${\bf k}_J(\vtheta) = [k_{\vpsi}(\vtheta_1,\vtheta),\dots,k_{\vpsi}(\vtheta_J,\vtheta)]^\top$, and $({\bf K})_{ij} = k_{\vpsi}(\vtheta_i,\vtheta_j)$ is the kernel matrix.  The GP posterior on $L(\vtheta)$ implies a probabilistic model over the  unnormalized posterior $p(\vtheta) = e^{L(\vtheta)}g(\vtheta)$, allowing us to compute 
\begin{align*}
\widehat{p}(\vtheta) = \Exp[p(\vtheta)|\mathcal{D}] &= g(\vtheta) \exp\left( \mu_p(\vtheta) + \frac{1}{2}\sigma_p^2(\vtheta)\right), \\
\mbox{var}[p(\vtheta)|\mathcal{D}] &= g(\vtheta)^2\exp\left( 2\mu_p(\vtheta) + \sigma_p^2(\vtheta)\right)\exp\left(  \sigma_p^2(\vtheta) - 1\right).
\end{align*}
Hence, this probabilistic model over $p(\vtheta)$ can be used to build acquisition functions and sequentially select new design points $\vtheta^*_t$ for $t=1,\dots, T$. For instance, we can select the point where the posterior variance is maximum \cite{kandasamy2017query},
$$
\vtheta^*_{t} = \arg\max_{\vtheta'}\mbox{var}[p(\vtheta')|\mathcal{D}_{t-1}],
$$
or the point that minimizes the expected integrated variance \cite{jarvenpaa2021parallel},
$$
\vtheta^*_{t} = \arg\min_{\vtheta'} \Exp\left[ \left. \int \mbox{var}[p(\vtheta)|\mathcal{D}_{t-1}\cup\{\vtheta'\}]  d\vtheta  \right \vert \vtheta ',\mathcal{D} \right].
$$
}

\end{document}